\documentclass{article}

\usepackage[preprint]{neurips_2025}

\usepackage[utf8]{inputenc} 
\usepackage[T1]{fontenc}    
\usepackage{CJKutf8}
\usepackage{hyperref}       
\usepackage{url}            
\usepackage{xurl}
\usepackage{booktabs}       
\usepackage{amsfonts}       

\usepackage{svg}
\usepackage{nicefrac}       
\usepackage{microtype}      
\usepackage[table,xcdraw]{xcolor}         

\usepackage{subcaption}
\usepackage{wrapfig}
\usepackage{multirow}
\usepackage{makecell}
\usepackage{xspace}
\usepackage{enumitem}
\usepackage{amsmath}
\usepackage{listings}
\usepackage{algorithm}
\usepackage{algorithmic}
\usepackage{amsfonts,amssymb}
\usepackage{amsthm}
\usepackage{mathrsfs}
\usepackage{subcaption}
\usepackage{natbib}
\setcitestyle{numbers,square}
\usepackage{graphicx}       
\usepackage{cleveref}
\usepackage{fontawesome}
\usepackage{utfsym}
\usepackage[most]{tcolorbox}
\usepackage{siunitx} 
\usepackage{tabularx}
\usepackage{CJKutf8}
\usepackage{fancyvrb}
\usepackage{colortbl}
\usepackage{array}
\usepackage{threeparttable}
\usepackage{CJKutf8}
\usepackage{CJK}
\usepackage{booktabs}
\usepackage[table]{xcolor}
\usepackage{lipsum}
\usepackage{listings}
\definecolor{demphcolor}{RGB}{144,144,144}

\newcommand{\hhide}[1]{}
\newcommand{\hide}[1]{}

\definecolor{zhipublue}{HTML}{3859FF}

\newtcolorbox{promptbox}[1][]{
  breakable,
  title=#1,
  colback=gray!5,
  colframe=black,
  colbacktitle=gray!15,
  coltitle=black,
  fonttitle=\bfseries,
  bottomrule=1.5pt,
  toprule=1.5pt,
  leftrule=1pt,
  rightrule=1pt,
  arc=0pt,
  outer arc=0pt,
  enhanced,
  before upper={\parindent=1.5em} 
}

\theoremstyle{definition}

\title{GLM-5: from Vibe Coding to Agentic Engineering}

 \author{
{GLM-5 Team}
~\\\\
Zhipu AI ~\&~ Tsinghua University\\\\
(For the complete list of authors, please refer to the \hyperref[sec:contribution]{Contribution} section)
{}
}

\begin{document}

\maketitle

\vspace{-1.5em}
\begin{abstract}

\vspace{-0.5em}

We present GLM-5, a next-generation foundation model designed to transition the paradigm of vibe coding to agentic engineering.
 Building upon the agentic, reasoning, and coding (ARC) capabilities of its predecessor, GLM-5 adopts DSA to significantly reduce training and inference costs while maintaining long-context fidelity. To advance model alignment and autonomy, we implement a new asynchronous reinforcement learning infrastructure that drastically improves post-training efficiency by decoupling generation from training. Furthermore, we propose novel asynchronous agent RL algorithms that further improve RL quality, enabling the model to learn from complex, long-horizon interactions more effectively. Through these innovations, GLM-5 achieves state-of-the-art performance on major open benchmarks. Most critically, GLM-5 demonstrates unprecedented capability in real-world coding tasks, surpassing previous baselines in handling end-to-end software engineering challenges. 
Code, models, and more information are available at \url{https://github.com/zai-org/GLM-5}.


     

\end{abstract}

\begin{figure}[!h]
    \centering
    \includegraphics[width=\linewidth]{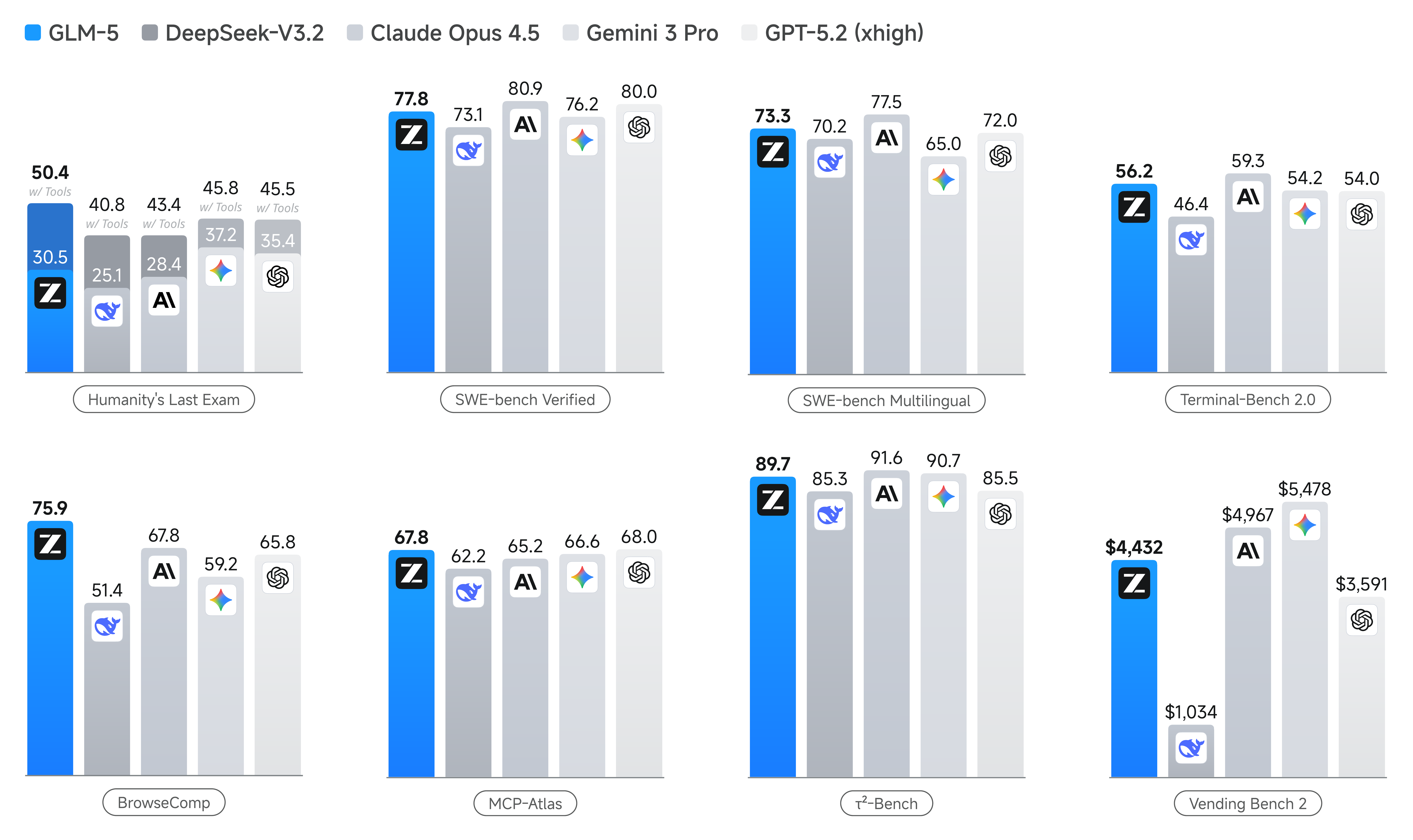}
    \caption{Results of GLM-5, DeepSeek-V3.2, Claude Opus 4.5, Gemini 3 Pro, and GPT-5.2 (xhigh) on 8 agentic, reasoning, and coding benchmarks: Humanity's Last Exam, SWE-bench Verified, SWE-bench Multilingual, Terminal-Bench 2.0, BrowseComp, MCP-Atlas, $\tau^2$-Bench, Vending Bench~2.}
    \label{fig:arc}
\end{figure}

\section{Introduction}

The pursuit of Artificial General Intelligence (AGI) requires not only scaling model parameters but also fundamentally rethinking the efficiency of intelligence and the architecture of autonomous improvement. With the release of GLM-4.5, we demonstrated that uniting Agentic, Reasoning, and Coding (ARC) capabilities into a single Model-of-Experts (MoE) architecture could yield state-of-the-art results across diverse benchmarks. However, as Large Language Models (LLMs) transition from passive knowledge repositories to active problem solvers, the dual challenges of computational cost and real-world adaptability—particularly in complex software engineering—have become the primary bottlenecks.

We present GLM-5, our next-generation flagship model designed to overcome these barriers. GLM-5 represents a paradigm shift in both performance and efficiency, achieving state-of-the-art status on major open leaderboards, including ArtificialAnalysis.ai, the LMArena Text, and the LMArena Code. More significantly, GLM-5 redefines the standard for real-world coding, demonstrating an unprecedented ability to handle complex, end-to-end software development tasks that go far beyond the scope of traditional static benchmarks like SWE-bench.

\paragraph{Results.} Figure~\ref{fig:arc} shows the results of GLM-5, GLM-4.7, Claude Opus 4.5, Gemini 3 Pro, and GPT-5.2 (xhigh) on 8 agentic, reasoning, and coding benchmarks: Humanity's Last Exam~\cite{phan2025humanity}, SWE-bench Verified~\cite{jimenez2023swe}, SWE-bench Multilingual~\cite{yang2025swemulti}, Terminal-Bench 2.0~\cite{tbench_2025}, BrowseComp~\cite{wei2025browsecomp}, MCP-Atlas~\cite{bandi2026mcp}, $\tau^2$-Bench~\cite{yao2024tau,barres2025tau2}, Vending Bench 2~\cite{backlund2025vending}. On average, GLM-5 achieves about 20\% improvement over our last version GLM-4.7, and is comparable to Claude Opus 4.5 and GPT-5.2 (xhigh), and better than Gemini 3 Pro.

\begin{figure}[t]
    \centering
    \includegraphics[width=1\linewidth]{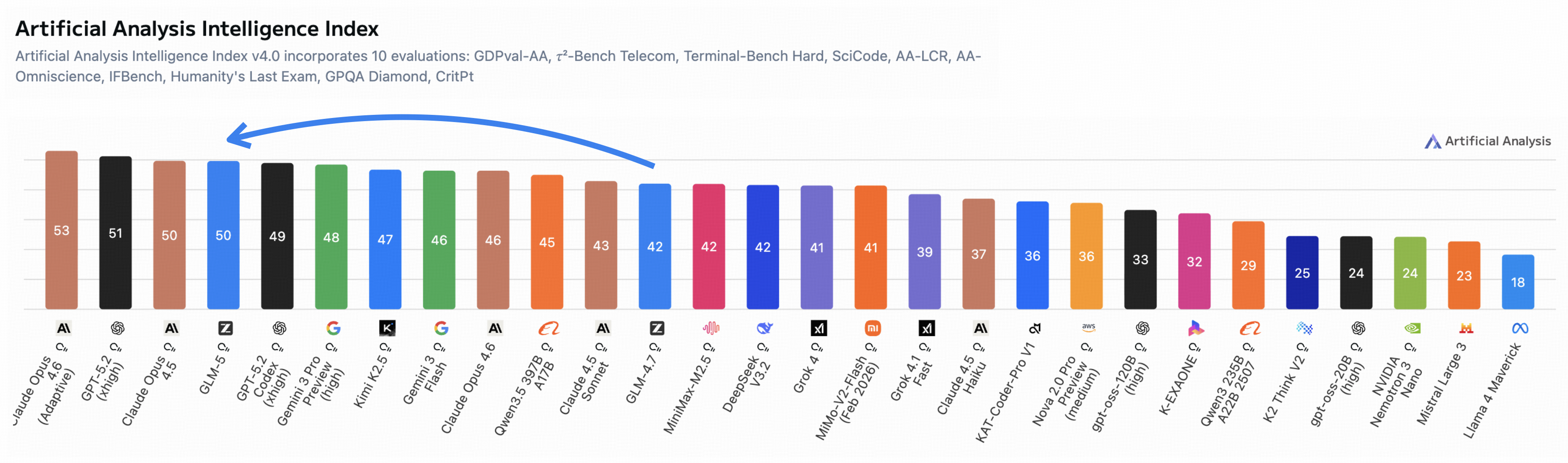}
    \caption{Artificial Analysis Intelligence Index v4.0 incorporates 10 evaluations: GDPval-AA, $\tau^2$-Bench Telecom, Terminal-Bench Hard, SciCode, AA-LCR, AA-Omniscience, IFBench, Humanity's Last Exam, GPQA Diamond, CritPt.}
    \label{fig:aa}
\end{figure}

GLM-5 scores 50 on the Intelligence Index v4.0 and is the new open weights leader (Cf. Figure~\ref{fig:aa}), up from GLM-4.7's score of 42 - an 8 point jump driven by improvements across agentic performance and knowledge/hallucination. This is the first time an open weights model has achieved a score of 50 on the Artificial Analysis Intelligence Index v4.0.

\begin{figure}[t]
    \centering
    \includegraphics[width=0.48\linewidth]{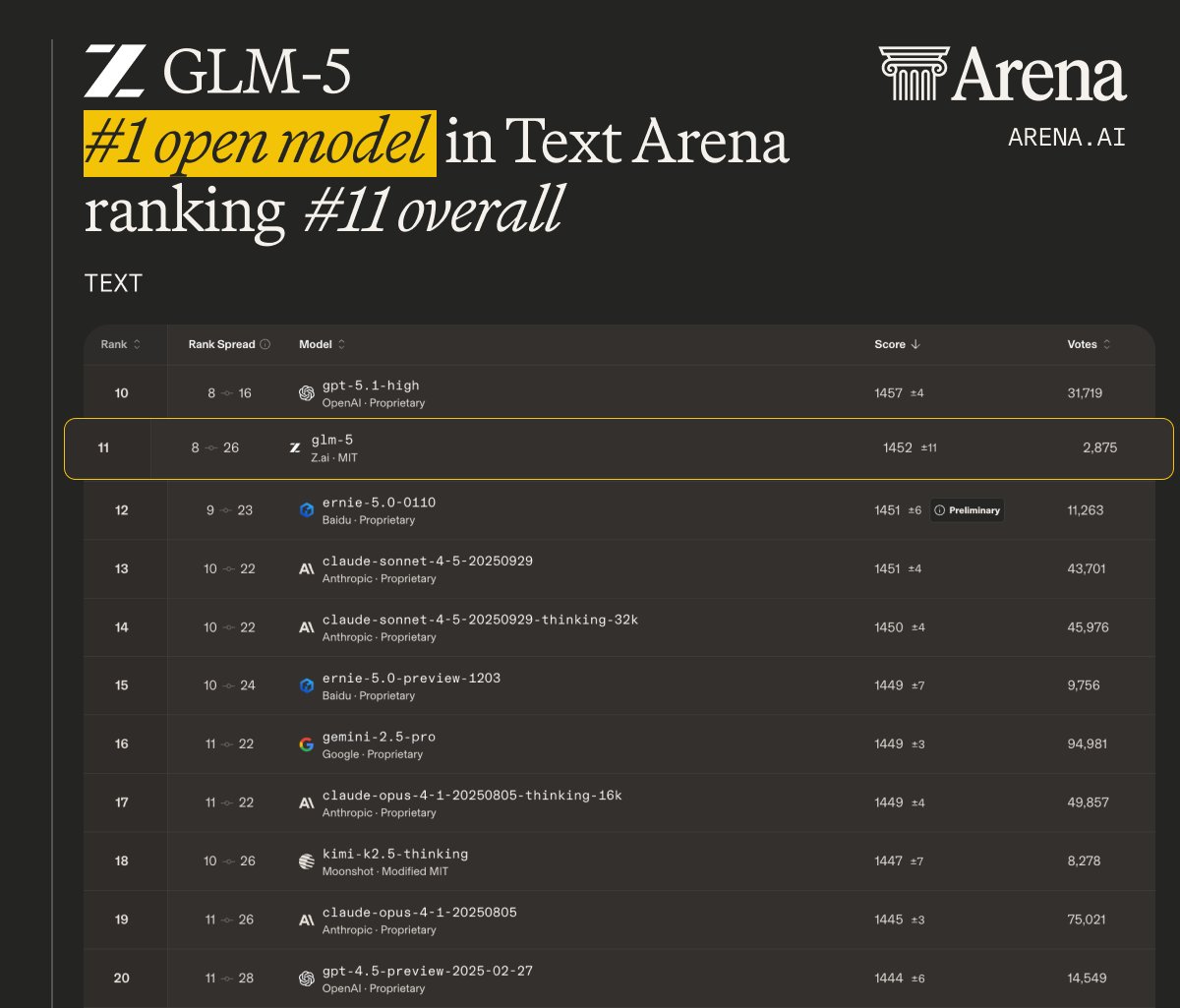} \;\;\;
    \includegraphics[width=0.48\linewidth]{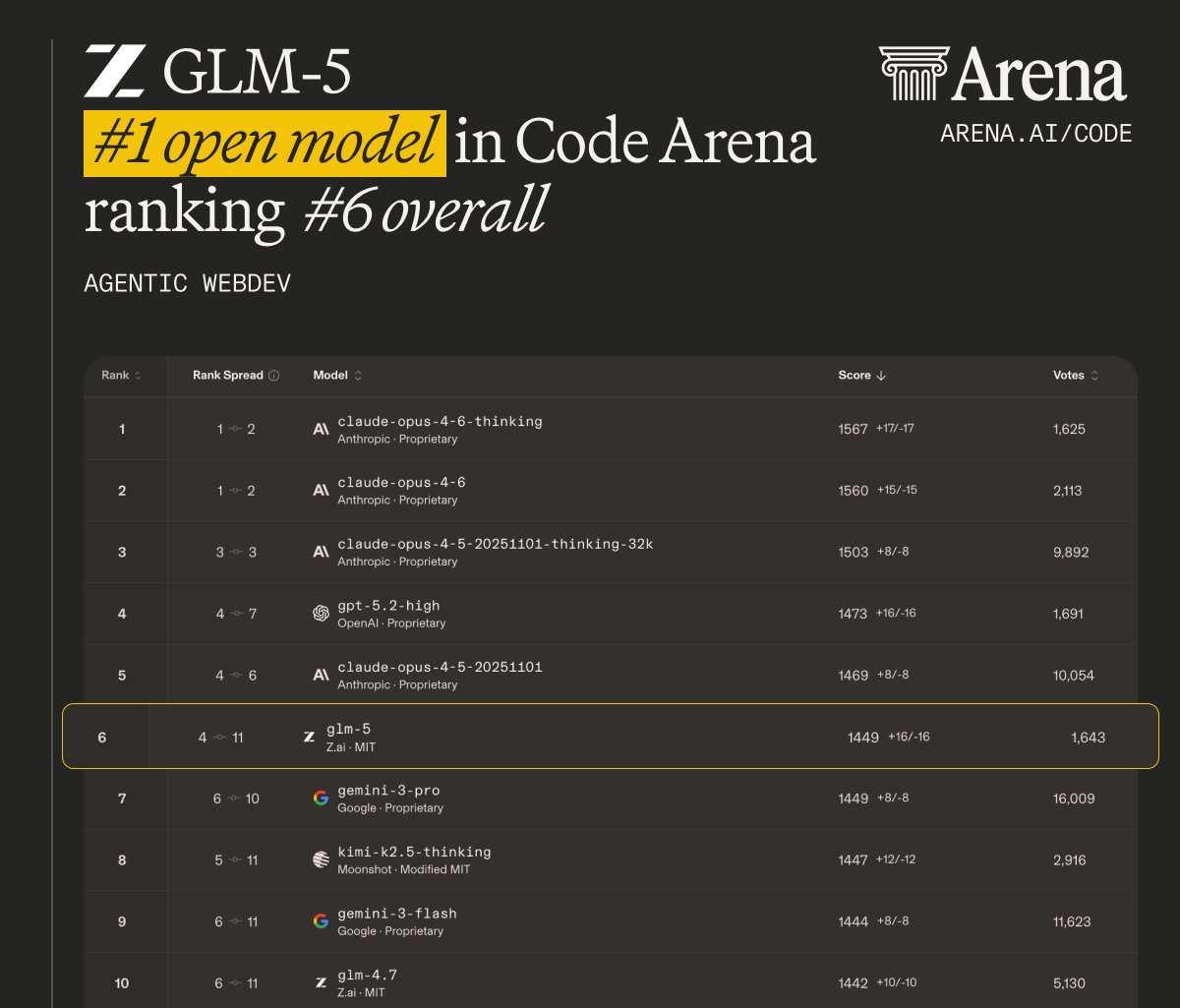}
    \caption{On LMArena, GLM-5 is the \#1 open model in both Text Arena and Code Arena.}
    \label{fig:arena}
\end{figure}

\begin{figure}[t]
    \centering
    \includegraphics[width=0.50\linewidth]{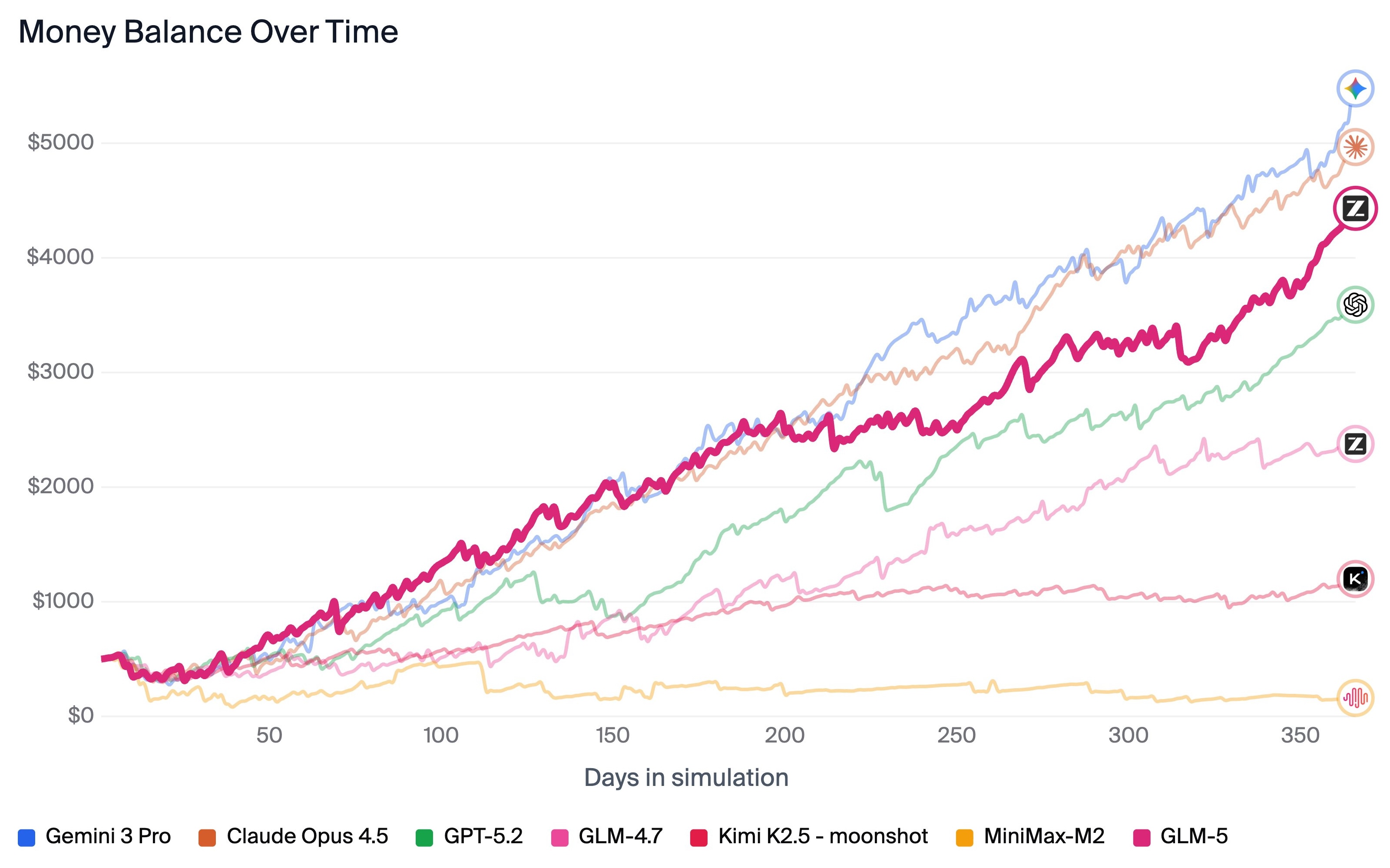} \;\;
    \includegraphics[width=0.47\linewidth]{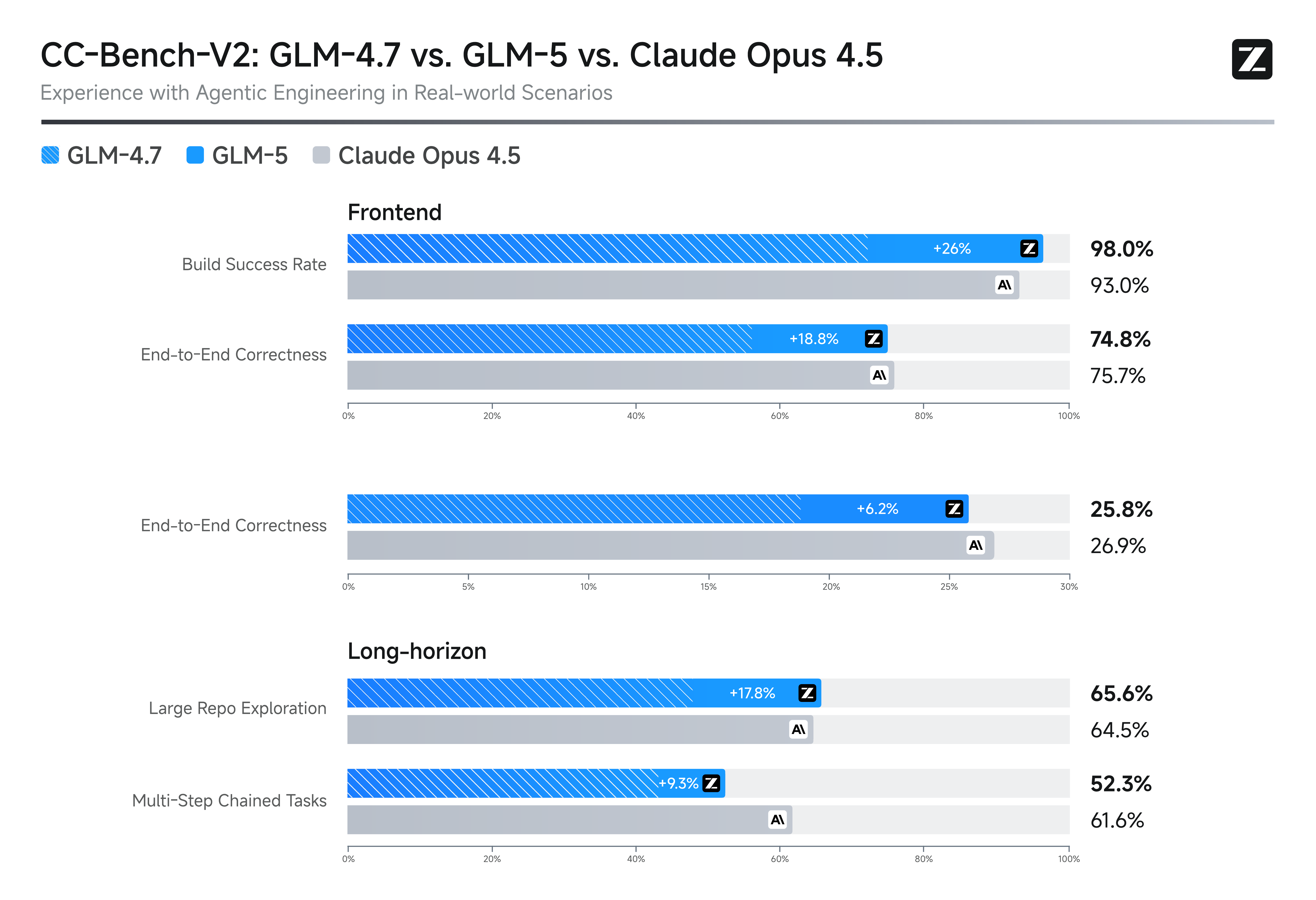}
    \caption{Results on several long-horizon tasks. Left: Vending-Bench 2; Right: CC-Bench-V2.}
    \label{fig:vendingcc}
\end{figure}

LMArena, initiated by UC Berkeley, is a transparent, shared space to evaluate and compare frontier AI capabilities by human judgment with millions of real tasks, including writing, coding, reasoning, designing, searching, and creating. The large volume of human interactions generates signals of real-world utility, making it different from the other static benchmarks.
Figure~\ref{fig:arena} shows that GLM-5 again is the \#1 open model in both Text Arena and Code Arena, and overall 
on par with Claude-Opus-4.5 and Gemini-3-pro.

Long-term coherence in agents becomes more and more important. Coding agents can now write code autonomously for hours, and the length and breadth of tasks AI models are able to complete are likely to increase. We use two benchmarks, Vending-Bench 2 and CC-Bench-V2, to evaluate how GLM-5 is able to complete long-horizon tasks.
Vending-Bench 2 is a benchmark for measuring AI model performance in running a business over long time horizons. Models are tasked with running a simulated vending machine business over a year and are scored on their bank account balance at the end.
Figure~\ref{fig:vendingcc} (left) shows that GLM-5 ranks \#1 among all open-source models, finishing with a final account balance of \$4,432. It approaches Claude Opus 4.5, demonstrating strong long-term planning and resource management.
Figure~\ref{fig:vendingcc} (right) further shows results on our internal evaluation suite CC-Bench-V2. GLM-5 significantly outperforms GLM-4.7 across frontend, backend, and long-horizon tasks, narrowing the gap with Claude Opus 4.5.

\begin{figure}[t]
    \centering
    \includegraphics[width=1\linewidth]{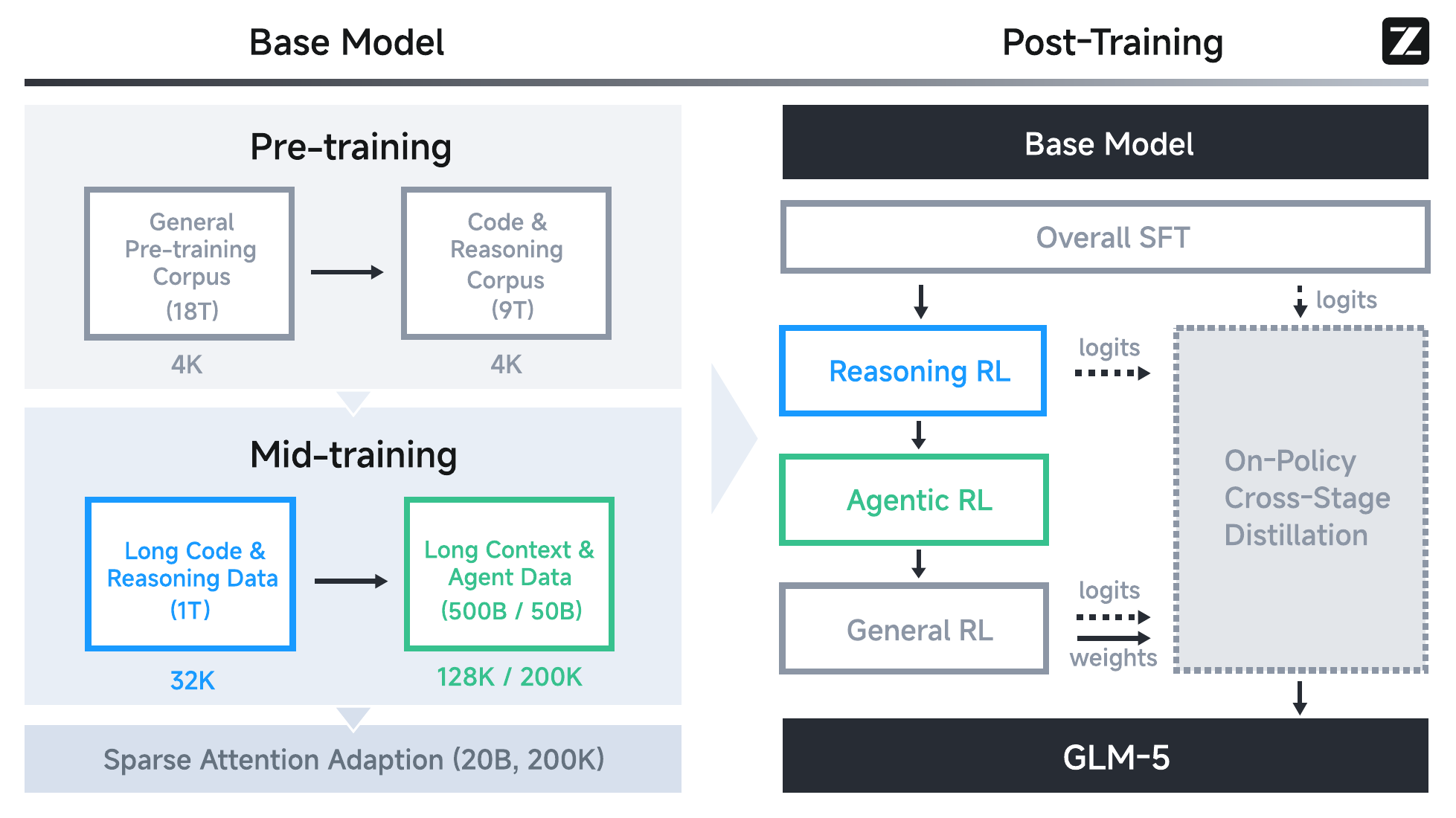}
    \caption{Overall training pipeline of GLM-5.}
    \label{fig:overall_pipeline}
\end{figure}

\paragraph{Methods.} Figure~\ref{fig:overall_pipeline} shows the overall training pipeline of GLM-5.
Our Base Model training began with a massive 27 trillion token corpus, prioritizing code and reasoning early on. We then employed a distinct Mid-training phase to progressively extend context length from 4K to 200K, focusing specifically on long-context agentic data to ensure stability in complex workflows.
In Post-Training, we moved beyond standard SFT. We implemented a sequential Reinforcement Learning pipeline—starting with Reasoning RL, followed by Agentic RL, and finishing with General RL. Crucially, we utilized On-Policy Cross-Stage Distillation throughout this process to prevent catastrophic forgetting, ensuring the model retains its sharp reasoning edge while becoming a robust generalist.
In summary, the leap in GLM-5’s performance is driven by the following technical contributions:

First, we adopt DSA (DeepSeek Sparse Attention)~\cite{deepseekai2025deepseekv32pushingfrontieropen}, a novel architectural innovation that significantly reduces both training and inference costs. While GLM-4.5 improved efficiency through a standard MoE architecture, DSA allows GLM-5 to dynamically allocate attention resources based on token importance, drastically lowering the computational overhead without compromising long-context understanding or reasoning depth. With DSA, we scale the model parameters up to 744B and extend the training token budget to ~28.5T tokens.

Second, we have engineered a new asynchronous reinforcement learning infrastructure. Building on the ``slime'' framework and the decoupled rollout engines initialized in GLM-4.5, our new infrastructure further decouples generation from training to maximize GPU utilization. This system allows for massive-scale exploration of agent trajectories without the synchronization bottlenecks that previously hampered iteration speed, significantly improving the efficiency of our RL post-training pipeline.

Third, we present novel asynchronous Agent RL algorithms designed to enhance the quality of autonomous decision-making. In GLM-4.5, we utilized iterative self-distillation and outcome supervision to train agents. For GLM-5, we have developed asynchronous algorithms that allow the model to learn from diverse, long-horizon interactions continuously. These algorithms are specifically optimized to improve the model's planning and self-correction capabilities in dynamic environments, directly contributing to our dominance in real-world coding scenarios.

Last, one more technical contribution lies in the fact that, from the first day, GLM-5 is full-stack adapted to Chinese GPU ecosystems. We have successfully completed deep optimization—spanning from underlying kernels to upper-level inference frameworks—across seven mainstream domestic chip platforms, including Huawei Ascend, Moore Threads, Hygon, Cambricon, Kunlunxin, MetaX, and Enflame.

With these advancements, GLM-5 stands not just as a more powerful model but as a more efficient and practical foundation for the next generation of AI agents. We release GLM-5 to the community to further advance the frontier of efficient, agentic general intelligence.

\section{Pre-Training}
Similar to GLM-4.5, the base model of GLM-5 goes through two stages: pre-training for general language and coding capacity, and mid-training for agentic and long-context capacity. We extend the training token budget for all the training stages of GLM-5, totaling 28.5 trillion tokens for the base model.

\subsection{Architecture}

\paragraph{Model size scaling.} GLM-5 scales to 256 experts and reduces its layer count to 80 to minimize expert parallelism communication overhead. This results in a 744B parameter model (40B active parameters), doubling the total size of GLM-4.5, which utilized 355B total and 32B active parameters.

\begin{table}[!ht]
    \small
    \centering
    \caption{Evaluation results for GQA-8 and variants of MLA.}
    \begin{tabular}{c|cccccccc}
        \toprule
        Dataset & Hellaswag & MMLU & C-Eval & RACE & BBH & GSM8K & HumanEval \\
        \midrule
        GQA-8 & 77.3 & 61.2 & 60.0 & 79.6 & \textbf{53.3} & \textbf{47.6} & \textbf{38.5} \\
        MLA & 77.3 & 61.5 & 59.7 & 77.8 & 48.9 & 46.2 & 33.5 \\
        MLA + Muon Split & \textbf{77.8} & \textbf{62.5} & \textbf{62.1} & \textbf{79.9} & 51.8 & 45.0 & 36.7\\
        MLA-256 + Muon Split & 77.4 & 62.0 & 59.9 & 79.6 & 51.3 & 47.5 & 36.6 \\
        \bottomrule
    \end{tabular}
    \label{tab:mla}
\end{table}

\paragraph{Multi-latent Attention.} By employing reduced key-value vectors, Multi-latent attention (MLA)~\cite{liu2024deepseekv2} matches the effectiveness of Grouped-Query Attention (GQA) but offers superior GPU memory savings and faster processing for long-context sequences.

However, in our experiments with Muon optimizer, we find that MLA with a 576-dimension latent KV-cache cannot match the performance of GQA with 8 query groups (denoted as GQA-8, 2048-dimension KV-cache). To overcome the performance gap, we propose an adaptation to the recipe of Muon optimizer in GLM-4.5.
In the original recipe, we apply matrix orthogonalization to the up-projection matrices $W^{UQ}, W^{UK}, W^{UV}$ for multi-head queries, keys, and values. Instead, we split these matrices into smaller matrices for different heads and apply matrix orthogonalization to these independent matrices. The method, denoted as Muon Split, enables projection weights for different attention heads to update at different scales. As shown in \Cref{tab:mla}, the method effectively improves the performance of MLA to match that of GQA-8. In practice, we also find that with Muon Split, the scale of attention logits of GLM-5 remains stable during pre-training without any clipping strategy.

Another disadvantage of MLA is its high computational cost during decoding. In decoding, MLA performs a 576-dimensional dot product, higher than the 128-dimensional computation of GQA. While the number of attention heads in DeepSeek-V3 is selected according to the roofline of H800~\cite{zhao2025insights}, it is inappropriate for other hardware. Given the Multi-head Attention (MHA) style of MLA during training and prefilling, we increase the head dimension from 192 to 256 and decrease the number of attention heads by 1/3. This keeps the training computation
and the number of parameters constant while decreasing the decoding computation. The variant, denoted as MLA-256 in \Cref{tab:mla}, matches the performance of MLA under Muon Split.

\begin{wraptable}{r}{0.4\linewidth}
    \centering
    \caption{Comparison of accept lengths of DeepSeek-V3.2 and GLM-5.}
    \label{tab:mtp}
    \begin{tabular}{c|c}
        \toprule
        Model  & Accept Length \\
        \midrule
        DeepSeek-V3.2 & 2.55 \\
        GLM-5  & \textbf{2.76}\\
        \bottomrule
    \end{tabular}
\end{wraptable}

\paragraph{Multi-token Prediction with Parameter Sharing.} Multi-token prediction (MTP)~\cite{gloeckle2024mtp,liu2024deepseek} increases the performance of base models and acts as draft models for speculative decoding~\cite{leviathan2023spec}. However, during training, to predict the next $n$ tokens, $n$ MTP layers are required. As a result, the memory usage of MTP parameters and the kv cache scales linearly with the number of speculative steps. Instead, DeepSeek-V3 is trained with a single MTP layer and predicts the next 2 tokens during inference. The training-inference discrepancy reduces the acceptance rate of the second token. Therefore, we propose sharing the parameters of 3 MTP layers during training. This keeps the memory cost of the draft model consistent with DeepSeek-V3 while increasing the acceptance rate. In \Cref{tab:mtp}, we show that the acceptance length of GLM-5 is longer than DeepSeek-V3.2, given the same number of speculative steps (4) in our private prompt set.

\subsubsection{Continued Pre-Training with DeepSeek Sparse Attention (DSA)}
\begin{table}[!ht]
    \centering
    \caption{Comparison of long-context benchmarks between MLA and DSA base models.}
    \begin{tabular}{c|cccc}
    \toprule
     & MQ-NIAH-128k & MV-NIAH-128k & SQuAD-128k & HotpotQA-128k \\
    \midrule
    MLA & \textbf{100.0} & 95.5 & 79.7 & \textbf{66.3} \\
    DSA & \textbf{100.0} & \textbf{97.0} & \textbf{86.0} & 63.0\\
    \bottomrule
    \end{tabular}
    \label{tab:dsa}
\end{table}
We use DSA in our training. 
The core philosophy of DSA~\cite{deepseekai2025deepseekv32pushingfrontieropen} is to replace the traditional dense $O(L^2)$ attention—which becomes prohibitively expensive at $128\text{K}$ contexts—with a dynamic, fine-grained selection mechanism. Unlike fixed patterns (like sliding windows), DSA ``looks'' at the content to decide which tokens are important.
What makes DSA particularly interesting from a researcher's perspective is how it was introduced via Continued Pre-Training from a dense base model. This avoided the ``astronomical'' cost of training from scratch. The transition follows a two-stage ``dense warm-up and sparse training adaptation'' strategy. DeepSeek-V3.2-Exp maintains the same benchmark performance as its dense predecessor, proving that ~90\% of attention entries in long contexts are indeed redundant.
DSA reduces the attention computation by roughly 1.5-2× for long sequences, which is very important for the reasoning-heavy agents we are building, being able to handle 128K contexts at half the GPU cost.

\begin{figure}
    \centering
    \includegraphics[width=0.6\linewidth]{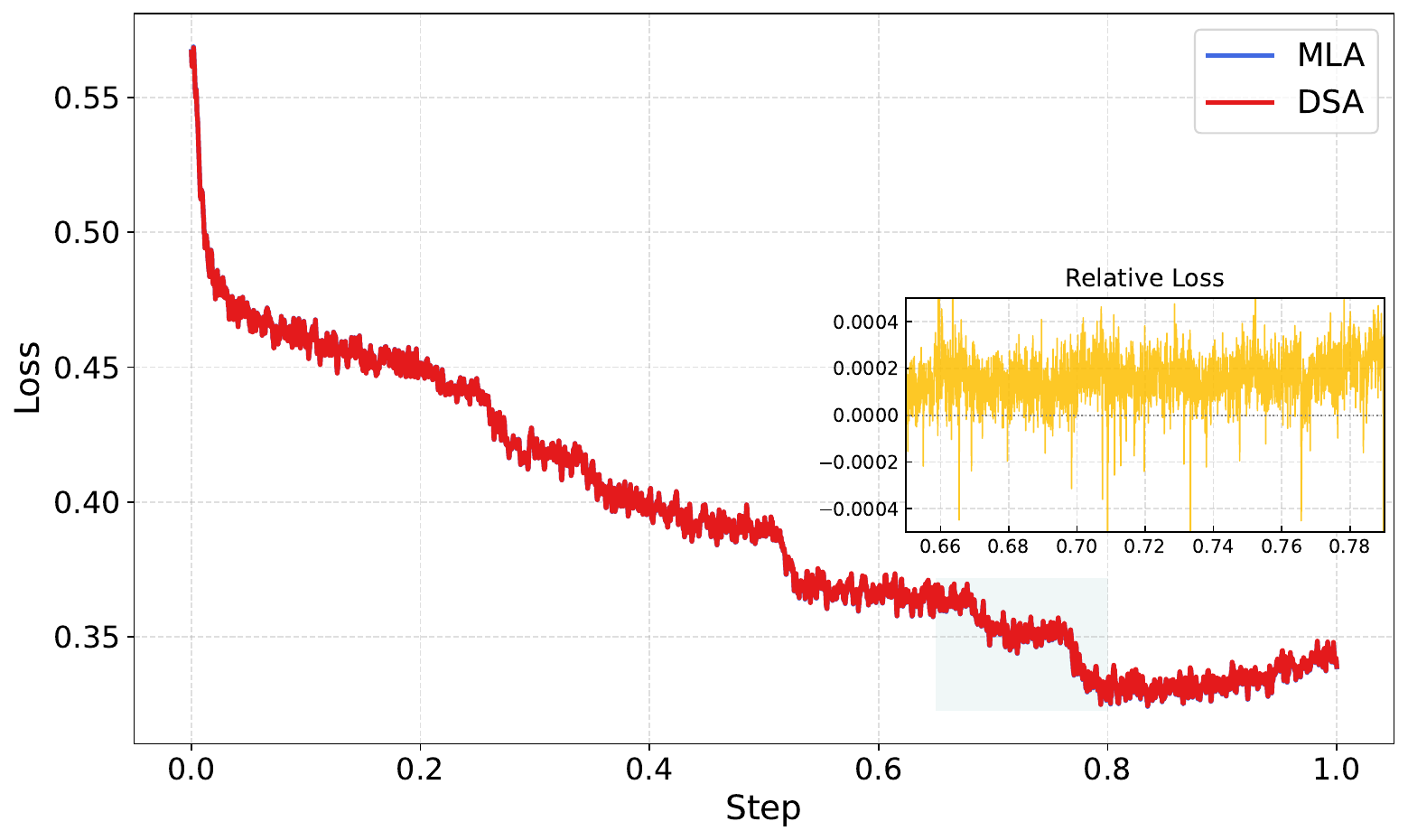}
    \caption{SFT loss curves comparison between MLA and DSA training. Results are smoothed by Running Average with a window size of 50.}
    \label{fig:placeholder}
\end{figure}

The DSA training begins from the base model at the end of mid-training. The warm-up stage goes through 1000 steps with each step trained on 14 sequences of 202,752 tokens and a maximum learning rate of 5e-3. The sparse adaptation stage follows the training data and hyperparameters of mid-training and goes through 20B tokens. Although the training budget is much smaller than that of DeepSeek-V3.2 (943.7B tokens), we find that it is enough to adapt the DSA model to match the performance of the original MLA model. As shown in \Cref{tab:dsa}, the long-context performance of the DSA model is close to that of the MLA model. To further validate the effectiveness of DSA training, we fine-tune the DSA and MLA models with the same SFT data, respectively, and find that the two models tie in training loss and evaluation benchmarks.

\subsubsection{Ablation Study of Efficient Attention Variants}

Beyond DSA~\cite{liu2025deepseek}, we explore several alternative efficient attention mechanisms based on GLM-9B\footnote{One of our GLM-4 series models, available at https://github.com/zai-org/GLM-4}. The baseline employs group query attention across all 40 layers and has been fine-tuned with a 128K-token context window. We evaluate the following approaches:

\begin{itemize}[leftmargin=1.5em,itemsep=2pt]
    \item {{Sliding Window Attention (SWA) Interleave:}} A fixed alternating pattern of
    full-attention and windowed-attention layers applied uniformly across the network.
    \item {{Gated DeltaNet (GDN)}~\cite{yanggated}}: A linear attention variant that
    replaces the quadratic softmax attention computation with a gated linear recurrence,
    reducing the computational cost of attention from quadratic to linear in sequence length.
\end{itemize}

\noindent Building on these baselines, we propose two improvements:

\begin{itemize}[leftmargin=1.5em,itemsep=2pt]
    \item {SWA Pattern (Search-Based):} Inspired by PostNAS~\cite{gu2025jet}, we introduce a search-based adaptation method that identifies the optimal subset of layers for SWA conversion while retaining full attention in the remaining layers. We employ a beam search strategy to determine the configuration that maximizes performance on long-context downstream tasks. To mitigate computational costs, we conduct the search exclusively at a 16K context length and generalize the resulting pattern to all other input lengths. Specifically, we use a beam size of~8, optimizing two layers per step; for GLM-9B (40 layers), the process converges in approximately 10 steps. At each step, candidate patterns are evaluated on the RULER benchmark~\cite{hsiehruler} at 16K context length, and the top-8 candidates are retained for the subsequent step. The final derived pattern is \texttt{SFSSFFSSSFFFFSSFSFFFFFFSFSFSSFSSFSFSSFSSS}, where \texttt{S} and \texttt{F} denote SWA and full-attention layers, respectively. As shown in Table~\ref{tab:ruler_results}, this search-based configuration significantly outperforms the fixed interleaved approach. Notably, despite being optimized only at 16K, the pattern exhibits robust length generalization, maintaining effective across all tested context lengths.

    \item {SimpleGDN:} A minimalist linearization strategy designed for maximal reuse of pre-trained weights, improving upon GDN for continual-training adaptation. We remove the \texttt{Conv1d} and explicit gating modules entirely and instead directly map the pre-trained Query, Key, and Value projection weights into the linear recurrence formulation. This simplification eliminates the need for additional parameters while preserving the efficiency benefits of linear attention.
\end{itemize}

\begin{table}[h]
    \small
    \centering
    \caption{RULER benchmark results for the GLM-9B baseline and two SWA variants \emph{without any additional training}. Both SWA methods use a 1:1 ratio of full-attention to SWA layers with a 4096-token window size. The search-based SWA pattern is discovered once at 16k context length and applied uniformly across all input lengths.}
    \label{tab:ruler_results}
    \begin{tabular}{lcccccc}
        \toprule
         & {4K} & {8K} & {16K} & {32K} & {64K} & {128K} \\
        \midrule
        GLM-9B (Full Attn)  & 95.19 & \textbf{93.67} & \textbf{92.01} & \textbf{91.09} & \textbf{85.35} & \textbf{75.28} \\
        SWA Interleave      & 94.87 & 54.02 & 25.89 & 12.61 &  8.32 &  6.51 \\
        SWA Pattern & \textbf{95.78} & 92.54 & 88.92 & 82.52 & 70.23 & 53.95 \\
        \bottomrule
    \end{tabular}
\end{table}

We evaluate all methods on four long-context benchmarks: RULER~\cite{hsiehruler}, MRCR\footnote{\url{https://huggingface.co/datasets/openai/mrcr}}, HELMET-ICL~\cite{yen2024helmet}, and RepoQA~\cite{liu2024repoqa}. Results are summarized in Table~\ref{tab:gqa_results_appendix}. We continually train each method on 190B tokens with a 64K context length, maintaining a 1:1 ratio between efficient attention layers and full attention layers. For the GDN and SimpleGDN methods, we follow the Jet-Nemotron~\cite{gu2025jet} pipeline.

\begin{table*}[h]
\centering
\caption{Long-context benchmark results. All efficient attention variants are continual-trained from the GLM-9B full-attention baseline. SWA pattern denotes search-based layer selection; SWA interleave denotes the fixed alternating pattern. $\Delta$@64K and $\Delta$@128K show the difference relative to the full-attention baseline at 64K and 128K context lengths, respectively.}
\label{tab:gqa_results_appendix}
\resizebox{\textwidth}{!}{
\begin{tabular}{lcccc}
\toprule
  & \textbf{RULER (64K\,/\,128K)}
  & \textbf{MRCR (64K\,/\,128K)}
  & \textbf{HELMET-ICL (64K\,/\,128K)}
  & \textbf{RepoQA (64K\,/\,128K)} \\
\midrule
GLM-9B 
  & 85.35\,/\,75.28
  & 36.53\,/\,35.39
  & 77.68\,/\,77.36
  & 69.00\,/\,65.83 \\
\midrule
SWA Interleave
  & 65.94\,/\,44.93 {\scriptsize($\downarrow$19.41\,/\,$\downarrow$30.35)}
  & 30.03\,/\,28.83 {\scriptsize($\downarrow$6.50\,/\,$\downarrow$6.56)}
  & 75.96\,/\,63.52 {\scriptsize($\downarrow$1.72\,/\,$\downarrow$13.84)}
  & 50.33\,/\,39.33 {\scriptsize($\downarrow$18.67\,/\,$\downarrow$26.50)} \\
SWA Pattern
  & \textbf{83.72\,/\,69.59} {\scriptsize($\downarrow$1.63\,/\,$\downarrow$5.69)}
  & \textbf{35.02\,/\,33.58} {\scriptsize($\downarrow$1.51\,/\,$\downarrow$1.81)}
  & 76.48\,/\,74.60 {\scriptsize($\downarrow$1.20\,/\,$\downarrow$2.76)}
  & 62.33\,/\,51.17 {\scriptsize($\downarrow$6.67\,/\,$\downarrow$14.66)} \\
GDN
  & 76.76\,/\,64.00 {\scriptsize($\downarrow$8.59\,/\,$\downarrow$11.28)}
  & 31.72\,/\,30.22 {\scriptsize($\downarrow$4.81\,/\,$\downarrow$5.17)}
  & 76.88\,/\,74.84 {\scriptsize($\downarrow$0.80\,/\,$\downarrow$2.52)}
  & 65.50\,/\,56.17 {\scriptsize($\downarrow$3.50\,/\,$\downarrow$9.66)} \\
SimpleGDN
  & 81.76\,/\,67.03 {\scriptsize($\downarrow$3.59\,/\,$\downarrow$8.25)}
  & 33.03\,/\,31.27 {\scriptsize($\downarrow$3.50\,/\,$\downarrow$4.12)}
  & \textbf{79.80\,/\,81.84} {\scriptsize($\uparrow$2.12\,/\,$\uparrow$4.48)}
  & \textbf{65.50\,/\,58.50} {\scriptsize($\downarrow$3.50\,/\,$\downarrow$7.33)} \\
\bottomrule
\end{tabular}
}
\end{table*}

The results in Table~\ref{tab:gqa_results_appendix} reveal a clear trade-off hierarchy among efficient attention methods. Naively interleaved sliding window attention (SWA) causes catastrophic degradation on long-context tasks (e.g., $-$30.35 on RULER@128K), while search-based layer selection substantially narrows this gap by preserving full attention where it matters most. Linear attention variants such as GDN further improve quality but at the cost of additional parameters; SimpleGDN strikes the best balance by maximally reusing pre-trained weights. Nevertheless, all of these methods incur an inherent accuracy gap on fine-grained retrieval tasks—up to 5.69 points on RULER@128K and 7.33 on RepoQA@128K—due to the unavoidable information loss introduced by efficient attention mechanisms during continual-training adaptation, even when half of the layers retain full attention. In contrast, DSA is lossless by construction: its lightning indexer achieves token-level sparsity without discarding any long-range dependencies, enabling application to all layers with no quality degradation.

To verify this, we conduct a small-scale DSA experiment on GLM-4.7-Flash\footnote{\url{https://huggingface.co/zai-org/GLM-4.7-Flash}} with multi-latent attention. Following the standard DSA recipe, training proceeds in two stages: (i)~a {warmup} phase that trains only the indexer for 1{,}000 steps (batch size 16) while keeping all base-model weights frozen, followed by (ii)~a {joint-training} phase in which both the model and the indexer are co-trained on 150B tokens.
Table~\ref{tab:dsa_ruler_results} summarizes the results on RULER across context lengths from 4K to 128K. Even the warmup-only variant ({GLM-4.7-Flash + DSA warmup}) already preserves the vast majority of baseline performance; the drop is modest and concentrated at the longest context window (128K: $79.21 \to 71.35$), while shorter contexts remain virtually unaffected. After the full 150B-token joint-training phase, {GLM-4.7-Flash + DSA} closes nearly all of this residual gap: it {surpasses} the baseline at 16K ($+0.86$), 32K ($+0.49$), and 64K ($+1.72$), while incurring only a 0.35-point deficit at 128K.

\begin{table}[t]
    \small
    \centering
    \caption{RULER benchmark results for the GLM-4.7-Flash with DSA.
    The warmup-only variant trains only the indexer while
    keeping the base model frozen, the full DSA variant jointly trains both for
    150B tokens. }
   \label{tab:dsa_ruler_results}
    \begin{tabular}{lcccccc}
        \toprule
         & {4K} & {8K} & {16K} & {32K} & {64K} & {128K} \\
        \midrule
        GLM-4.7-Flash  & 97.44 & 96.72 & 95.83 & 92.96 & 85.34 & 79.21 \\
        GLM-4.7-Flash + DSA warmup & 97.51 & 96.54 & 95.40 & 90.09 & 84.05 & 71.35  \\
        GLM-4.7-Flash + DSA & 96.77 & 96.25 & 96.69 & 93.45 & 87.06 & 78.86 \\
        \bottomrule
    \end{tabular}
\end{table}

\subsection{Pre-training Data}

\paragraph{Web.} Building upon the GLM-4.5 data pipeline, we refined our selection criteria for massive web datasets. We introduced another DCLM~\cite{li2025datacomplmsearchgenerationtraining} classifier based on sentence embeddings to identify and aggregate additional high-quality data beyond standard classifiers. To address the challenge of long-tail knowledge, we utilized a World Knowledge classifier—optimized via Wikipedia entries and LLM-labeled data—to distill valuable information from otherwise medium-low-quality data.

\paragraph{Code.} We expand the code pre-training corpus with refreshed snapshots from major code hosting platforms and a larger collection of code-containing web pages, resulting in a 28\% increase in fuzzily deduplicated unique tokens. To improve corpus integrity and reduce noise, we fix metadata alignment issues in Software Heritage code files and adopt a more accurate language classification pipeline. We follow GLM-4.5’s quality-aware sampling strategy for source code and code-related web documents. In addition, we train dedicated classifiers for a broader set of low-resource programming languages (e.g., Scala, Swift, Lua, etc.), improving sampling quality for these languages.

\paragraph{Math \& Science.} We collect high-quality math \& science data from webpages, books, and papers to further increase the reasoning abilities. Specifically, the content extraction pipelines for webpages and PDF parsing mechanisms
for books and papers are refined to increase data quality. We adopt large language models to score candidate documents and only retain the most educational content. For long-context documents, we develop a chunk-and-aggregate scoring algorithm to increase scoring accuracy. Filtering pipelines are conducted to strictly avoid the use of synthetic, AI-generated, or template-based data.

\subsection{Mid-Training}

Building upon the mid-training framework introduced in GLM-4.5, we scale up both the training volume and the maximum context length in GLM-5 to further strengthen the model's reasoning, long-context, and agentic capabilities.

\paragraph{Extended context and training scale.} We progressively extend the context window across three stages: 32K (1T tokens), 128K (500B tokens), and 200K (50B tokens). Compared to the 128K maximum in GLM-4.5, the additional 200K stage substantially improves the model's ability to process ultra-long documents and complex multi-file codebases. Long documents and synthetic agent trajectories are up-sampled at the later stages accordingly.

\paragraph{Software engineering data.} We retain the paradigm of concatenating repo-level code files, commit diffs, GitHub issues, pull requests, and relevant source files into unified training sequences. In GLM-5, we relax the repository-level filtering criteria to broaden the pool of eligible repositories, yielding approximately 10 million issue–PR pairs, while strengthening quality filtering at the individual issue level to reduce noise. We also retrieve a larger set of relevant files for each issue–PR pair, resulting in richer development contexts and broader coverage of real-world software engineering scenarios. After filtering, the issue–PR portion of the dataset comprises approximately 160B unique tokens.

\paragraph{Long-context data.} Our long-context training set comprises both natural and synthetic data. Natural data is curated from books, academic papers, and documents from general pre-training corpora employing multi-stage filtering (PPL, deduplication, length) and upsampling knowledge-intensive domains. In synthetic data construction, inspired by NextLong\cite{gao2025nextlongeffectivelongcontexttraining} and EntropyLong\cite{jia2025entropylongeffectivelongcontexttraining}, we employed diverse techniques to build long-range dependencies. Highly similar texts were aggregated via interleaved packing to produce sequences, aiming to mitigate the lost-in-the-middle phenomenon and improve performance across a range of long-context tasks. At the 200K stage, we additionally incorporated a small proportion of MRCR-like data, with multiple variants designed to extend OpenAI's original paradigm, to strengthen recall in extended multi-turn dialogues. Empirically, we find that increasing data diversity progressively enhances the model's long-context performance; notably, a subsequent 200K mid-training stage, building upon the initial 128K phase, further bolstered the model’s performance even within the 128K context window.

\subsection{Training Infrastructure}

\subsubsection{Memory Efficiency}

\textbf{Flexible MTP placement.} Under interleaved pipeline parallelism~\cite{narayanan2021efficientlargescalelanguagemodel}, model components are flexibly assigned to stages. The MTP module spans embedding, transformer, and output components. It incurs substantially higher memory usage than other modules, leading to stage-level imbalance. We co-locate the MTP output layer with the main output layer on the final stage to enable parameter sharing, while placing its embedding and transformer components on the preceding stage. This reduces memory pressure on the final stage and improves balance across pipeline ranks.

\textbf{Pipeline ZeRO2 gradient sharding.} Each pipeline rank maintains multiple stages~\cite{narayanan2021efficientlargescalelanguagemodel}, and naively each stage requires a full gradient buffer for accumulation and optimizer updates. Inspired by ZeRO2~\cite{rajbhandari2020zeromemoryoptimizationstraining}, we shard gradients across data-parallel ranks so that each stage stores only a 1/dp fraction of the full gradients. In addition, we retain full accumulation buffers for only two stages at a time and reuse them via double buffering. While one stage buffer accumulates gradients over consecutive microbatches, gradient synchronization for the previous stage buffer is performed in parallel. This reduces persistent gradient memory to per-stage sharded buffers plus only two full buffers for rolling accumulation, without additional synchronization overhead in practice.

\textbf{Zero-redundant communication for the Muon distributed optimizer.} Naive Muon implementations all-gather full model parameters on each data-parallel rank, causing transient memory spikes and redundant communication. We restrict all-gather to parameter shards owned by each rank and overlap local computation with shard communication. This eliminates redundant communication and significantly reduces optimizer-related peak memory overhead.

\textbf{Pipeline activation offloading.} During pipeline warmup, forward execution advances ahead of backpropagation, prolonging the lifetime of intermediate activations. We offload the activations to host memory after forward execution and reload them prior to backward execution~\cite{298555}. Offloading is applied at layer granularity to further reduce peak memory usage. Combined with fine-grained recomputation, this largely eliminates the need to keep activations resident in GPU memory. Offload and reload are scheduled to overlap with computation while avoiding contention with peer-to-peer communication and MoE token routing (dispatch and combination). This substantially reduces the activation memory footprint with near-zero overhead.

\textbf{Sequence-chunked output projection for peak memory reduction.} 
Output projection and cross-entropy loss incur transient memory overhead from storing activations for backpropagation and promoting them to higher precision during loss computation. To reduce this overhead, we partition the input sequence into smaller chunks and compute projection and loss independently on each chunk, completing forward and backward passes and releasing activations before moving on. As a result, peak memory usage decreases as the number of chunks increases. With an appropriate chunk count, this approach alleviates output-layer memory pressure while maintaining performance comparable to unchunked execution.

\subsubsection{Parallelism Efficiency}

\textbf{Efficient deferred weight gradient computation.} To reduce pipeline bubbles, we defer some weight gradient computation of the critical path~\cite{qi2023zero}. Fine-grained deferral with optimized storage and communication overlap improves throughput while keeping memory overhead bounded.

\textbf{Efficient long-sequence training.} Longer sequences exacerbate load imbalance across data parallel and pipeline parallel groups. We address this through workload-aware sequence reordering, dynamic redistribution of attention computation, and flexible partitioning of data parallel ranks into context-parallel groups of varying sizes~\cite{ge2025bytescale,wang2025flexsp}. A hierarchical all-to-all overlaps intra-node and inter-node communication for QKV tensors to reduce latency.

\subsubsection{INT4 Quantization-aware training}

To provide better accuracy at low-precision, we apply INT4 QAT in the SFT stage. Moreover, to further mitigate the training time overhead, we have developed a quantization kernel applicable to both training and offline weight quantization, which ensures bitwise-identical behavior between training and inference.

\section{Post-Training}

The post-training phase of GLM-5 aims to transform the base model into a highly capable assistant with robust reasoning, coding, and agentic abilities. As illustrated in Figure~\ref{fig:overall_pipeline}, our pipeline follows a progressive alignment strategy: starting with multi-task Supervised Fine-Tuning (SFT) that introduces sophisticated interleaved thinking modes, followed by specialized Reinforcement Learning (RL) stages for reasoning and agentic tasks, and concluding with a general RL stage for human-style alignment. By leveraging on-policy cross-stage distillation as the final refinement, GLM-5 effectively mitigates capability regression while harnessing the performance gains from each training stage.

\subsection{Supervised Fine-Tuning}

\begin{figure}[!t]
    \centering
    \includegraphics[width=\linewidth]{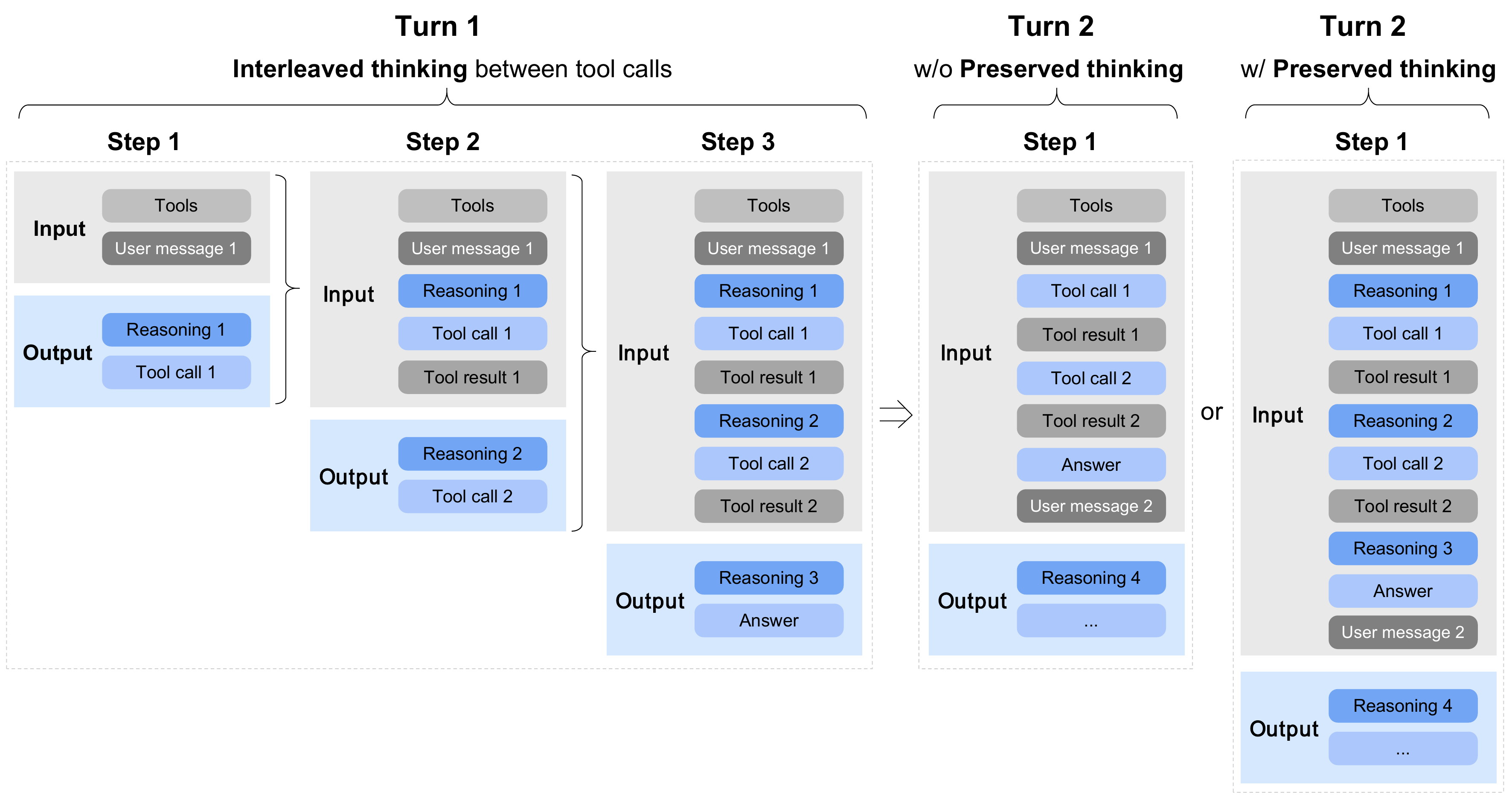}
    \caption{Illustration of Interleaved Thinking and Preserved Thinking.}
    \label{fig:thinking_mode}
\end{figure}

Compared with GLM-4.5, GLM-5 significantly expands the scale of \emph{Agent} and \emph{Coding} data during the SFT stage. The SFT corpus of GLM-5 covers three major categories:
\begin{itemize}[leftmargin=1.5em,itemsep=2pt]
    \item \textbf{General Chat}: question answering, writing, role-playing, translation, multi-turn dialogue, and long-context interactions;
    \item \textbf{Reasoning}: mathematical, programming, and scientific reasoning;
    \item \textbf{Coding \& Agent}: frontend and backend engineering code, tool calling, coding agents, search agents, and general-purpose agents.
\end{itemize}

Additionally, GLM-5 extends the maximum context length to 202,752 tokens during SFT. Along with an updated chat template, the model supports three distinct thinking characteristics (see Figure \ref{fig:thinking_mode}), including:

\begin{itemize}[leftmargin=1.5em,itemsep=2pt]
    \item \textbf{Interleaved Thinking}: the model thinks before every response and tool call, improving instruction following and the quality of generation\footnote{Interleaved thinking was first introduced by \url{https://platform.claude.com/docs/en/build-with-claude/extended-thinking\#interleaved-thinking}}.
    \item \textbf{Preserved Thinking}: in coding agent scenarios, the model automatically retains all thinking blocks across multi-turn conversations, reusing existing reasoning instead of re-deriving it from scratch. This reduces information loss and inconsistencies, and is well-suited for long-horizon, complex tasks\footnote{Preserved thinking was also adopted by Claude since Opus 4.5. See \url{https://platform.claude.com/docs/en/build-with-claude/extended-thinking\#thinking-block-preservation-in-claude-opus-4-5-and-later}}.
    \item \textbf{Turn-level Thinking}: the model supports per-turn control over reasoning within a session—disable thinking for lightweight requests to reduce latency/cost, enable it for complex tasks to improve accuracy and stability.
\end{itemize}

By thinking between actions and maintaining consistency across turns, GLM-5 achieves more stable and controllable behavior on complex tasks.

For \textbf{General Chat}, we optimize the response style to be more logical and concise compared to GLM-4.5. For role-playing tasks, we collect and construct a broader and more diverse dataset covering multiple languages and role configurations. In particular, we define several evaluation dimensions---including instruction following, linguistic expressiveness, creativity, logical coherence, and long-dialogue consistency---and apply both automatic and human filtering to curate and refine the data.

For \textbf{Reasoning} tasks, we further enhance the depth of the model's reasoning. Specifically, for logical reasoning, we construct verifiable problems and synthesize high-quality data using rejection sampling. For mathematical and scientific problems, a difficulty-based filtering process is applied, retaining only problems that are challenging for the GLM-4.7 model.

For \textbf{Coding} and \textbf{Agent} tasks, compared to GLM-4.5, GLM-5 constructs a large number of execution environments to obtain high-quality trajectories, with particular emphasis on real-world scenarios and long-horizon tasks. We further improve the SFT data using expert reinforcement learning and rejection sampling. Erroneous segments within trajectories are retained but masked out in the loss function, allowing the model to learn error correction behaviors without reinforcing incorrect actions.

\subsection{Reasoning RL}

\paragraph{RL algorithm backbone.}
Our RL algorithm builds upon GRPO~\cite{shao2024deepseekmath} and incorporates the IcePop technique~\cite{IcePop2025} to mitigate the \emph{training-inference mismatch}, i.e., the discrepancy between the inference distribution and the training distribution during RL optimization. We explicitly distinguish between the \emph{training policy} $\pi^{\text{train}}$, used for gradient updates, and the \emph{inference policy} $\pi^{\text{infer}}$, used for trajectory sampling. Compared to the original IcePop formulation, we remove the KL regularization term to accelerate RL improvement.
The final optimization loss is:
\begin{equation}
\begin{aligned}
\mathcal{L}(\theta)=
-\mathbb{E}_{
x \sim \mathcal{D},
\{y_i\}_{i=1}^{G} \sim \pi^{\text{infer}}_{\theta_{\text{old}}}(\cdot \mid x)
}
&\Bigg[
\frac{1}{G}
\sum_{i=1}^{G}
\frac{1}{|y_i|}
\sum_{t=1}^{|y_i|}
\operatorname{pop}
(\rho_{i,t}, 1/\beta, \beta) \\
&\cdot
\min\!\left(
r_{i,t}\hat{A}_{i,t},
\operatorname{clip}\!\left(
r_{i,t},
1-\epsilon_{\text{low}},
1+\epsilon_{\text{high}}
\right)
\hat{A}_{i,t}
\right)
\Bigg],
\label{icepop}
\end{aligned}
\end{equation}
where the training-inference mismatch ratio is defined as
\begin{align*}
\rho_{i,t}
=
\frac{
\pi_{\theta_{\text{old}}}^{\text{train}}(y_{i,t} \mid x, y_{i,<t})
}{
\pi_{\theta_{\text{old}}}^{\text{infer}}(y_{i,t} \mid x, y_{i,<t})
}.
\end{align*}

The operator $\operatorname{pop}(\cdot)$ suppresses samples whose mismatch ratio deviates excessively:
\begin{align*}
\operatorname{pop}(\rho_{i,t}, 1/\beta, \beta)
=
\begin{cases}
\rho_{i,t}, & 1/\beta \le \rho_{i,t} \le \beta, \\
0, & \text{otherwise}.
\end{cases}
\end{align*}

The PPO-style importance ratio and the group-normalized advantage follow the original GRPO definition:
\begin{align*}
r_{i,t}
=
\frac{
\pi_\theta^{\text{train}}(y_{i,t}\mid x,y_{i,<t})
}{
\pi_{\theta_{\text{old}}}^{\text{train}}(y_{i,t}\mid x,y_{i,<t})
},
\quad
\hat{A}_{i,t}
=
\frac{
R_i - \operatorname{mean}(R_1,\dots,R_G)
}{
\operatorname{std}(R_1,\dots,R_G)
}.
\end{align*}

During training, we set hyperparameters $\beta=2, \epsilon_{\text{low}}=0.2, \epsilon_{\text{high}}=0.28$. Training is performed entirely on-policy with a group size of 32 and a batch size of 32.

\paragraph{DSA RL insights.}
We conduct a very large-scale RL training on a model based on the DSA architecture. Compared with MLA, DSA introduces an additional indexer that retrieves the top-k most relevant key-value entries and computes attention sparsely over the retrieved subset.
The retrieved top-k results are critical for RL stability. This is analogous to how MoE models use routing replay~\cite{zheng2025group} to preserve the activated top-k experts to ensure training-inference consistency.
However, directly adapting this strategy to indexer replay, i.e., storing the indexer’s top-k indices at every token position is clearly impractical, since the $k = 2048$ used by the indexer is much larger than the $k$ typically used in MoE, and storing all these indices would incur enormous storage costs as well as significant communication overhead between the training engine and the inference engine.

We find that adopting a deterministic top-k operator effectively resolves the training-inference mismatch in DSA indexer token selection. Compared with the non-deterministic CUDA-based top‑k implementation used in SGLang’s DSA Indexer, directly using the naive \texttt{torch.topk} is slightly slower but deterministic. It produces more consistent outputs and yields substantial RL gains. In contrast, other non-deterministic top‑k operators (e.g., CUDA or TileLang implementations) caused drastic performance degradation during RL after only a few steps, accompanied by a sharp drop in entropy. Therefore, throughout our RL stages, we use \texttt{torch.topk} as the default top-k operator in the DSA Indexer in our training engine. We also freeze the indexer parameters by default during RL to accelerate training and prevent unstable learning in the indexer.

\paragraph{Mixed domain reasoning RL.} In the Reasoning RL stage, we perform mixed RL training over four domains: mathematics, science, code, and tool-integrated reasoning (TIR). For mathematics and science, we curate data from both open-source datasets~\cite{du2025nemotron, moshkov2025aimo} and co-developed collections with external annotation vendors. We further apply difficulty filtering to focus training on problems that GLM-4.7 solves correctly only rarely or fails consistently, while remaining solvable by stronger teacher models (e.g., GPT-5.2 xhigh and Gemini 3 Pro Preview). For code, we cover both competitive programming style tasks and scientific coding tasks. The former is primarily sourced from Codeforces and representative datasets such as TACO~\cite{li2023taco} and SYNTHETIC-2-RL~\cite{synthetic2_blog_2025}, while the latter is constructed from internal problem pools by decomposing questions into the minimal code implementations required for correct solutions. For TIR, we reuse the more challenging subset of mathematics and science RL data, and additionally co-build STEM questions with annotation vendors that are explicitly designed to be answered with external tools. During RL training, we assign domain and source-specific judge models or evaluation systems to produce binary outcome rewards. We keep the overall mixture roughly balanced across the four domains, and consistently observe stable and significant gains in each domain under the mixed RL setting.

\subsection{Agentic RL}
To facilitate agentic performance of GLM-5, we develop a fully asynchronous and decoupled RL framework and optimize GLM-5 in coding and search agent tasks. Naive synchronous RL suffers from severe GPU idle time during long-horizon agent rollouts. By decoupling inference and training engines via a central Multi-Task Rollout Orchestrator, we achieve high-throughput joint training across diverse agentic workloads.

To maintain training stability under asynchronous off-policy conditions, we introduce two key mechanisms. First, a Token-in-Token-out (TITO) gateway eliminates re-tokenization mismatches by preserving exact action-level correspondence. Second, we employ a Direct Double-sided Importance Sampling, which applies a token-level clipping mechanism ($[1-\epsilon_\ell, 1+\epsilon_h]$) to rollout log-probabilities, while efficiently controlling off-policy bias without tracking historical policy checkpoints. We also employ a DP-aware routing to maximize KV-cache reuse during long-context inference for large-scale MoE models for speed up. To scaling agentic environments, we scale verifiable training environments across three domains: over 10K real-world Software Engineering (SWE), terminal tasks, and high-difficulty multi-hop search tasks. More details about agentic RL can be found in the subsequent Section 4.

\subsection{General RL}

\paragraph{Multi-dimensional optimization objectives.}
We decompose the optimization objectives of General RL into three complementary dimensions: \emph{foundational correctness}, \emph{emotional intelligence}, and \emph{task-specific quality}.

The \emph{foundational correctness} dimension serves as the bedrock of response quality. It targets a broad spectrum of error types that undermine the usability of model outputs, including instruction-following failures, logical inconsistencies, factual inaccuracies, knowledge hallucinations, and language disfluencies. The goal is to minimize the error rate so that responses reach a \emph{usable} baseline. We consider this a prerequisite for all subsequent optimization: a response containing factual errors or misinterpreting the user's intent can actively mislead the user, no matter how polished it may appear.

The \emph{emotional intelligence} dimension optimizes user experience beyond core correctness. It aims to produce responses that are empathetic, insightful, and stylistically close to natural human communication, making interactions with the model feel more natural and engaging.

The \emph{task-specific quality} dimension targets fine-grained optimization across various specific tasks. Building on the usability established by foundational correctness, it aims to elevate responses from merely correct to genuinely high-quality within each task category. This dimension covers a wide range of tasks, including writing, text processing, subjective and objective question answering, role-playing, and translation. Each task domain demands distinct reward signals, necessitating a hybrid reward system.

\paragraph{Hybrid reward system.}
To supervise the diverse objectives above, we build a hybrid reward system that integrates three complementary types of reward signals: \emph{rule-based reward functions}, \emph{outcome reward models} (ORMs), and \emph{generative reward models} (GRMs). Each has distinct strengths and weaknesses, and their combination is key to a stable, efficient, and scalable General RL training process.

Rule-based rewards provide precise and interpretable signals, but are limited to aspects expressible as deterministic rules. ORMs offer low-variance signals and high training efficiency, but are more susceptible to reward hacking, where the policy exploits superficial patterns rather than genuinely improving core capability. GRMs leverage language models to produce scalar or structured evaluations and are more robust to such exploitation, but tend to exhibit higher variance. By blending these three signal types, we obtain a reward system that balances precision, efficiency, and robustness, mitigating the weaknesses of any single component.

\paragraph{Human-in-the-loop style alignment.}
A distinctive aspect of our General RL pipeline is the explicit incorporation of high-quality human-authored responses. Rather than relying solely on model-generated responses, we introduce expert human responses as stylistic and qualitative anchors. This is motivated by the observation that purely model-generated optimization tends to converge toward recognizably ``model-like'' patterns---often verbose, formulaic, or lacking the nuance of skilled human writing. By exposing the model to human-written exemplars, we encourage it to adopt more natural, human-aligned response patterns.

\subsection{On-Policy Cross-Stage Distillation}
In our multi-stage RL pipeline, sequentially optimizing for distinct objectives can lead to the cumulative degradation of previously acquired capabilities. To mitigate this issue, we perform on-policy cross-stage distillation as the \textbf{final stage}, adopting an on-policy distillation algorithm~\cite{guminillm,qwen3,xiao2026mimov2flashtechnicalreport,lu2025onpolicydistillation} to swiftly recover the skills acquired in earlier SFT and RL stages (Reasoning RL and General RL). Specifically, the final checkpoints from the preceding training stages serve as teacher models, where the training prompts are sampled from the corresponding teachers' RL training sets and mixed in appropriate proportions. The training loss can be obtained by replacing the advantage term in Eq.~\ref{icepop} with the following formula (`sg' stands for the stop gradient operation, e.g., \texttt{.detach()}):

\begin{equation}
    \hat{A}_{i,t} = \text{sg}\left[\log\frac{\pi_{\theta_{\text{teacher}}}^{\text{infer}}(y_{i,t}\mid x,y_{i,<t})}{\pi_\theta^{\text{train}}(y_{i,t}\mid x,y_{i,<t})}\right].
\end{equation}

Currently, we utilize the inference engine to fetch teachers' logits. In the future, we plan to migrate the inference backend to the training engine and uniformly adopt the Multi-Query Attention (MQA) mode of MLA for inference ($\pi_{\theta_{\text{teacher}}}^{\text{infer}}\rightarrow\pi_{\theta_{\text{teacher}}}^{\text{train}}$).
During training, the group size in the GRPO algorithm is configured to 1 to increase data throughput, and the batch size is set to 1024. This is feasible at this stage because it is no longer necessary to maintain a large group of samples per prompt to estimate advantages; the advantage is computed directly from the gap with the teacher models instead.

\subsection{RL Training Infrastructure: The \emph{slime} Framework}

We continue to use slime as the unified post-training infrastructure for GLM-5, enabling end-to-end reinforcement learning (RL) at scale. Rather than introducing new system components, GLM-5 \emph{fully leverages} slime's capabilities to (1) broaden task coverage via free-form rollout customization and a server-based execution model, (2) substantially increase throughput via mixed-precision training/rollouts together with MTP and Prefill-Decode (PD) disaggregation---particularly for multi-turn RL workloads, and (3) improve robustness through heartbeat-driven rollout fault tolerance and router-level server lifecycle management.

\subsubsection{Scaling Out: Flexible Training via Highly Customizable Rollouts}
GLM-5’s post-training spans a diverse spectrum of objectives. To support this diversity without task-specific forks, GLM-5 leverages slime's highly customizable rollout interface together with its server-based rollout execution.

\textbf{Highly customizable rollouts.} slime provides a flexible interface for implementing task-specific rollout logic---including multi-turn interaction loops, tool invocation, environment feedback handling, and verifier-guided branching---without modifying the underlying infrastructure. GLM-5 leverages this capability to support a broad range of domains and training paradigms, including but not limited to reasoning RL, general RL, agentic RL, and on-policy distillation, all within a unified training stack.

\textbf{Server-based rollouts via HTTP APIs.} slime exposes its rollout servers and inference router through standard HTTP APIs, allowing users to interact with slime's serving layer in the same way as a conventional inference engine. This decouples rollout logic from the training process boundary: external agent frameworks and environments can call the server/router endpoints directly, while the optimization backend remains unchanged for both short-horizon single-turn training and long-horizon multi-turn trajectories.

\subsubsection{Scaling Up: Tail-Latency Optimization for RL Rollouts}
For RL rollouts, the optimization target is not aggregate throughput but \emph{end-to-end latency}, dominated by the slowest (long-tail) sample in each step. In practice, a single straggling trajectory can stall synchronization points (e.g., batch completion, buffer readiness, trainer updates) and directly determine wall-clock progress. GLM-5 therefore fully leverages slime's latency-oriented serving and scheduling mechanisms to minimize both median latency and, more importantly, tail latency.

\textbf{No-queue serving via multi-node inference with DP-attention for MLA.}
To avoid queueing delays, rollout requests must be served promptly even under bursty traffic, which requires substantial KV-cache capacity. GLM-5 adopts a multi-node inference deployment (e.g., EP64 and DP64 over 8 nodes) to provision sufficient distributed KV-cache. DP-attention is primarily introduced to prevent copying KV across different ranks.

\textbf{Tail-latency reduction with FP8 rollouts and MTP.}
GLM-5 uses FP8 for rollout inference to reduce per-token latency and shorten the completion time of long trajectories. In addition, GLM-5 leverages slime's support for Multi-Token Prediction (MTP), which is especially effective under the small-batch decoding regime typical in RL rollouts. Since tail latency is often driven by small-BS stragglers (e.g., rare long contexts, complex multi-turn reasoning, tool-heavy traces), MTP provides disproportionately large benefits on the long tail, improving the time-to-completion of the slowest sample and thus reducing step-level stall time.

\textbf{PD disaggregation to prevent prefill-decode interference in multi-turn RL.}
In multi-turn settings, long-prefix prefills are frequent (conversation history, tool traces, code context). Under DP-attention, mixing prefill and decode on the same serving resources can create severe interference: a heavy prefill can preempt or disrupt ongoing decodes on the server, preventing other samples from making continuous progress and sharply worsening tail latency. GLM-5, therefore, leverages slime's Prefill--Decode (PD) disaggregation. By running prefills and decodes on dedicated resources, decodes remain stable and uninterrupted, enabling long-horizon samples to progress continuously and significantly improving tail behavior in multi-turn agentic RL.

\subsubsection{Rollout Robustness: Heartbeat-Driven Fault Tolerance}

At scale, transient failures (e.g., individual server crashes, network issues, or performance degradation) are inevitable. GLM-5 leverages slime's heartbeat-driven fault-tolerance to ensure training continuity under such events: rollout servers periodically emit heartbeats monitored by the orchestration layer, and unhealthy servers are proactively terminated and deregistered from the inference router. As a result, retries are automatically routed away from failed or degraded servers to healthy ones, preventing single-server incidents from interrupting rollouts and preserving uninterrupted end-to-end RL training.

\section{Agentic Engineering}
We describe the transition from \textbf{vibe coding} (human prompting) to \textbf{agentic engineering}.
In vibe coding, a human prompts an AI model to write code. In agentic engineering, AI agents write the code themselves. They plan, implement, and iterate. 
To support these long-horizon tasks, GLM-5 utilizes a fully asynchronous and decoupled RL framework to significantly boost GPU utilization by reducing idle time during agent rollouts. 
To scaling agent environments, we have developed environment-building pipelines. For coding tasks, we set up real-world  software engineering issues and terminal tasks by creating over 10,000 verifiable training scenarios. For search agents, we develop an automatic and scalable complex multi-step reasoning data synthesis pipeline to build data for agentic training.

\subsection{Asynchronous RL for Agentic Tasks}
To conduct RL for agent tasks, we design a fully asynchronous and decoupled RL infrastructure that efficiently handles long-horizon agent rollouts and supports flexible multi-task RL training across diverse agent frameworks.

We adopt the group-wise policy optimization algorithm for RL training.
For each problem \(x\), we sample \(K\) agent traces \(\{y_1,\dots,y_K\}\) from the previous policy
\(\pi_{\text{old}}\), and optimize the model \(\pi_\theta\) with respect to the following objective:
\[
L(\theta)
   = \mathbb{E}_{x\sim\mathcal{D}}\!\left[
        \frac{1}{K}\sum_{i=1}^{K}
        \left(
            r(x,y_i) - \bar{r}(x)
        \right)
     \right],
\]
where $\bar{r}(x) \;=\; \frac{1}{K}\sum_{i=1}^{K} r\bigl(x,y_i\bigr)$
is the mean reward of the sampled responses. It is noted that only model-generated tokens are used for optimization, and the environment feedback is ignored in loss computation. 

\subsubsection{Asynchronous RL Design for Agentic Training}

Due to the long-tail nature of the rollout process, naive synchronous RL training introduces substantial bubbles during the rollout stage because of the severely imbalanced generation of agentic tasks, which can cause large GPU idle time. 
To improve training throughput, we adopt a fully asynchronous training paradigm for Agentic RL to boost GPU utilization and training efficiency.
Concretely, we decouple the training engine and the inference engine onto different GPU devices. 
The inference engine continuously generates trajectories. Once the number of generated trajectories reaches a predefined threshold, the batch is sent to the training engine to update the model.
To reduce policy lag and keep the training approximately on-policy, the model weights used by the rollout engine are periodically synchronized with those of the training engine. The training engine updates the model parameters and pushes the new weights back to the inference engine every $K$ gradient updates.
While asynchrony could significantly improve overall training efficiency, it also means that different trajectories may be generated by different versions of the model, introducing a severe off-policy issue. 
Since the weight update considers a different optimization problem due to the changing
rollout policy, we also reset the optimizer after each weight update of the inference engine.

\paragraph{Server-based multi-task training design.}
To address the heterogeneity of trajectory generation in multi-task RL, where different tasks typically rely on distinct tool sets and task-specific rollout logic, we introduce a server-based Multi-Task Rollout Orchestrator for multi-task RL training.
This component is designed to ensure seamless compatibility between the slime RL training framework and diverse downstream tasks through a central orchestrator with multiple registered task services. Specifically, each task implements its own rollout and reward logic as an independent microservice, which is registered with the central orchestrator for management and scheduling.
During the rollout stage, the central orchestrator controls the per-task rollout ratio and generation speed to achieve balanced data collection across tasks. Crucially, we standardize trajectories from all agentic tasks into a unified message-list representation. This enables joint training of complex agentic frameworks (e.g., Software Engineering task) while also supporting centralized post-processing and logging for heterogeneous workloads. This design cleanly isolates task-specific logic from the core training loop, enabling seamless integration with multi-task RL training.
Serving as the backbone of the GLM-5 training infrastructure, this orchestrator supports over 1k concurrent rollouts and enables automated, dynamic adjustment of task sampling ratios, as well as fine-grained monitoring of task progress.

\subsubsection{Optimizing Asynchronous Training Stability}
\paragraph{Token-in-Token-out vs. Text-in-Text-out.}
In an RL rollout setting, \emph{token-in-token-out} (TITO) means the training pipeline consumes the \emph{exact} tokenization and decoded-token stream produced by the inference engine, and uses it directly to build trajectories for learning. 
In contrast, \emph{text-in-text-out} treats the rollout engine as a black box that returns finalized text; the trainer then reconstructs the trajectory by re-tokenizing that text (and often re-deriving boundaries and truncation) before computing losses. 
This seemingly small choice is consequential: re-tokenization can introduce subtle mismatches in token boundaries, whitespace/normalization handling, truncation, or special-token placement, which in turn can corrupt step alignment between actions and rewards/advantages—especially when rollouts are streamed, truncated, or interleaved across many actors. 
We find token-in-token-out is critical for asynchronous RL training because it preserves exact action-level correspondence between what was sampled and what is optimized while enabling actors to emit trajectory fragments (token IDs + metadata) immediately without a lossy text round-trip and without waiting for post-hoc re-tokenization on the learner side.
In practice, we implement a TITO Gateway that intercepts all generation requests from rollout tasks and records each trajectory’s token IDs and metadata. 
This design isolates the cumbersome token ID processing from downstream agent rollout logic, while avoiding re-tokenization mismatches during RL training.

\paragraph{Direct
double-sided importance sampling for token clipping.}
Unlike the synchronous RL training setting in Section 3, in the asynchronous setting, rollout engines may undergo multiple updates during a single trajectory generation,
which renders the tracking of exact behavior probabilities $\pi_{\theta_{\text{old}}}$ computationally prohibitive. Otherwise, we have to maintain an extensive history of model checkpoints $\{\pi_{\theta_{\text{old}}^{(1)}}, \dots, \pi_{\theta_{\text{old}}^{(N)}}\}$, which is infeasible in practical implementation.

To resolve this, we first employ a simplified token-level importance sampling mechanism that reuses the log-probabilities generated during rollout as a direct behavior proxy. By calculating the importance sampling ratio as $r_t(\theta) = \frac{\pi_{\theta}}{\pi_{\text{rollout}}}$ and discarding the traditional $\pi_{\theta_{\text{old}}}$, we eliminate the computational overhead of separate old-policy inference. Second, we employ a double-sided calibration token-level masking strategy. Instead of the asymmetric clipping used in standard PPO, we restrict the trust region to $[1-\epsilon_\ell, 1+\epsilon_h]$, where $\epsilon_\ell$ and $\epsilon_h$ are clipping hyperparameters. Tokens falling outside this interval are entirely masked from gradient computation to prevent instabilities caused by extreme policy divergence. This shares similarities with the IcePop mechanism \cite{team2025every}, yet our strategy is simpler by further removing the $\pi_{\theta_{\text{old}}}$ and achieving more stable training. 

Formally, the optimization objective with token-level clipping can be written as:
\begin{equation}
    L(\theta) = \mathbb{E}_t \left[ f(r_t(\theta), \epsilon_l, \epsilon_h) \hat{A}_t \log \pi_{\theta}(a_t|s_t) \right]
\end{equation}
In this formulation, the importance sampling ratio $r_t(\theta)$ is computed as:
\begin{equation}
    r_t(\theta) = \exp\left( \log \pi_\theta(a_t|s_t) - \log \pi_{\text{rollout}}(a_t|s_t) \right)
\end{equation}
Stability is further enforced via the calibration function $f(x; \epsilon_\ell, \epsilon_h)$:
\begin{equation}
    f(x; \epsilon_\ell, \epsilon_h) = 
    \begin{cases} 
    x, & \text{if } 1-\epsilon_\ell < x < 1+\epsilon_h \\
    0, & \text{otherwise}
    \end{cases}
\end{equation}

In the experiments, we find that reusing rollout log-probabilities accepts a controlled degree of off-policy bias to circumvent the need for historical policy tracking while boosting training stability.

\paragraph{Dropping off-policy and noisy samples.}
In asynchronous RL, overly long trajectories can become highly off-policy, which may destabilize training. To filter out these severely off-policy samples, we log the policy weight version used by the rollout engine at generation time. Specifically, for each response we record the sequence of model versions involved, $(w_0, \ldots, w_k)$ with $w_0 < \cdots < w_k$. Let $w'$ denote the current policy version. We discard a sample if its oldest rollout version is too stale, i.e., if $w' - w_0 > \tau$, where $\tau$ is a predefined threshold. This removes trajectories that lag too far behind the current policy.

Additionally, coding-agent sandboxes can be inherently unstable and may fail for reasons unrelated to the model (e.g., environment crashes). Such failures introduce noisy training signals because they reflect environment instability rather than the model’s capability. To mitigate this, we record the failure reason for each sample and exclude samples that fail due to environment collapse. For group-based sampling methods such as GRPO, removing failed samples can leave an incomplete group. In that case, we pad the group by repeating valid samples if the number of valid samples exceeds half of the group size; otherwise, we drop the entire group. This procedure reduces spurious reward noise and improves training stability.

\paragraph{DP-aware routing for acceleration.}
We propose a DP-aware routing mechanism to preserve KV cache locality under Data Parallelism (DP) for large-scale MoE inference. In multi-turn agentic workloads, sequential requests from the same rollout share an identical prefix. To maximize KV reuse, we enforce rollout-level affinity: all requests belonging to a given agent instance are routed to the same DP rank. Concretely, we introduce a stateful routing layer that maps each rollout ID to a fixed DP rank using consistent hashing. This mapping remains stable across turns, eliminating cross-rank cache misses. To prevent long-term imbalance, we combine hashing with lightweight dynamic load rebalancing over the hash space. This design avoids redundant prefill computation without requiring KV synchronization across DP ranks. As rollout length increases, prefill cost remains proportional to incremental tokens rather than total context length. The result is improved end-to-end latency and higher effective throughput for long-context agentic inference.

\subsection{Environment Scaling for Agents}

To support reinforcement learning across diverse agentic tasks, we construct verifiable, executable environments that provide grounded feedback for both code-centric and content-generation workflows.
For agentic coding tasks, we develop two environment-building pipelines that construct verifiable executable environments: an environment setup pipeline built upon real-world software engineering issues, and a synthesis pipeline for terminal-agent environments.
Beyond coding, we further introduce a slide generation environment, in which the agent operates over structured HTML with executable rendering and layout-based verification.

\subsubsection{Software Engineering (SWE) Environments} 
Before constructing executable environments, we collect a large corpus of real-world Issue-Pull Request (PR) pairs and apply rigorous rule-based and LLM-based filtering to ensure the acquisition of authentic, high-quality issue statements. We categorize these instances into different task types--bug fixing, feature implementation, refactoring, and others--and include the necessary task requirements to ensure that the model's implementation is consistent with the test patch. We employ an environment setup pipeline based on the RepoLaunch~\citep{zhang2025swebenchgoeslive} framework that scales the construction of executable environments from real-world SWE issues. This pipeline automatically analyzes a repository’s installation and dependency setup to build an executable environment and generate test commands, then leverages LLM to generate language-aware log-parsing functions from test outputs, enabling the extraction of  Fail-to-Pass (F2P) and Pass-to-Pass (P2P) test cases. Using this pipeline, we construct over 10k verifiable environments across thousands of repositories spanning 9 programming languages, including Python, Java, Go, C, CPP, JavaScript, TypeScript, PHP, and Ruby. 

\subsubsection{Terminal Environments} 

\paragraph{Synthesis from seed data.} To build verifiable terminal-agent environments at scale, we design an agentic data synthesis pipeline comprising three phases: task draft generation, concrete task implementation, and iterative task optimization. Starting from a set of seed tasks collected from real-world software engineering and terminal-based computer-use scenarios, we leveraged LLM to brainstorm and generate a large pool of verifiable terminal-task drafts. These drafts are then instantiated by a construction agent into concrete tasks in the Harbor \cite{harborframeworkteam2026harborframework} format, including structured task descriptions, Dockerized execution environments, and corresponding test scripts. Subsequently, a refine agent inspects and iteratively refines the generated tasks according to manually defined rubrics, ensuring that Docker images can be built reliably, test cases are consistent with task specifications, and the environments are robust against potential exploits or shortcuts. Overall, the pipeline yields thousands of diverse and verifiable terminal-agent environments with Docker construction accuracy exceeding 90\%.


\paragraph{Synthesis from web-corpus.}
We develop a scalable, automated pipeline and construct LLM-verified terminal-based coding tasks based on web corpus, using a closed-loop design where the constructing agent also serves as its own first-pass evaluator.
First, we collect a large-scale corpus of code-relevant web pages and apply a data quality classifier to retain only high-quality content, discarding pages that are predominantly non-technical or lack substantive code content. From the filtered subset, we further identify web pages amenable to terminal-style task formulation. We then apply stratified sampling across topic categories and difficulty levels to ensure distributional balance and diversity in the resulting task pool.
Second, we prompt a coding agent with the Harbor task construction specification\footnote{\url{https://harborframework.com/docs/tasks/task-tutorial}}, including the task schema, formatting requirements, and exemplar tasks, alongside each selected source web page. The agent is instructed to (i) synthesize a complete terminal task grounded in the web page content, and (ii) execute the Harbor validation script against its own output. Upon validation failure, the agent iteratively diagnoses and revises the task until it passes all automated checks. Only tasks that successfully clear this self-verification loop are admitted into the final dataset.

\subsubsection{Search Tasks}

For deep-search information-seeking tasks, we build a data-synthesis pipeline that produces challenging multi-hop QA pairs. Each question requires multi-step reasoning grounded in evidence aggregated from multiple web sources.

\textbf{Web Knowledge Graph (WKG) Construction and Question Generation.}
Starting from trajectories of an early-stage search agent, we collect and deduplicate all encountered URLs, retaining over two million high-information web pages across diverse domains. 
The LLM performs semantic parsing for entity recognition, noise filtering, and structured information extraction. 
The WKG is continuously updated with new pages and refined using downstream verification signals via entity alignment, attribute normalization, relation consolidation, and semantic-consistency corrections. 
Based on the WKG, we sample low- to mid-frequency entities as seed nodes and expand their multi-hop neighborhoods to form complete subgraphs, while controlling expansion to reduce overlap. 
Using prompts targeting high-difficulty, multi-domain reasoning, we convert each subgraph into a question that implicitly encodes multi-entity relational chains. 

\textbf{High-Difficulty Question Filtering and Verification.}
We apply a three-stage pipeline to balance difficulty and correctness:
(1) Remove questions that a tool-free reasoning model correctly answers in at least one of eight independent attempts.
(2) Filter out questions solvable by an early-stage agent with basic search, browsing, and computation within a few steps. 
(3) Apply a verification agent for bidirectional validation: we collect candidate answers from the search trajectories in stage 2, then independently verify the question–answer consistency for both the candidates and the annotated ground truth, rejecting samples with non-unique answers, inconsistent evidence, or incorrect labels. 
This yields high-quality, high-difficulty, reliable multi-hop QA pairs.

\subsubsection{Inference with Context Management for Search Agents}

We find that the performance on BrowseComp~\cite{wei2025browsecomp} is sensitive to both the judge prompt and the judge model, and open-source judges can introduce systematic bias. To ensure consistency and reproducibility, we standardize all judge-based components using the official OpenAI evaluation prompt and the proprietary model o3-mini as the judge. Our case studies indicate this configuration aligns best with human-annotated ground truth, so we adopt it for all search agent evaluations.

Prior work~\citep{liu2025deepseek} has introduced context management, where \textit{Discard-all} resets the context by removing the entire history of tool calls. We further observe that model accuracy degrades substantially under extremely long contexts (e.g., beyond 100k tokens). Motivated by this, we employ a simple \textit{Keep-recent-k}  strategy. When the interaction history exceeds a threshold $k$, the content older than the most recent $k$ rounds will be folded to control context length. 
Let the trajectory be
$(q,r_1,a_1,o_1,r_2,a_2,o_2,\cdots,r_n,a_n,o_n)$, 
where $q$ denotes the question, $r_i$ denotes the reasoning at round $i$, $a_i$ the action (we design \textit{search}, \textit{open}, \textit{find} and \textit{python} 4 tools), and $o_i$ the tool observation. We fold only observations earlier than the most recent $k$ rounds:
$
    o_i \leftarrow \text{Tool result is omitted to save tokens.}\quad i=1, \ldots, n-k
$.
In our experiments, we set $k=5$, which yields a stable improvement and improves GLM-5 from 55.3\%($w/o$ keep-recent-$k$) to 62.0\%($w/$ keep-recent-$k$). We also find that using different values of keep recent $k$ or alternatively triggering keep-recent once the context length reaches a predefined token threshold, leads to the same results.

Building on this, we combine keep-recent with Discard-all to form a hybrid \textit{Hierarchical Context Management} strategy. During inference with keep-recent, if the total context length exceeds a threshold $T$, we discard the entire tool-call history and restart with a fresh context, while continuing to apply the keep-recent strategy. We select $T=32k$ via parameter search.

\begin{figure}
    \centering
    \includegraphics[width=0.7\linewidth]{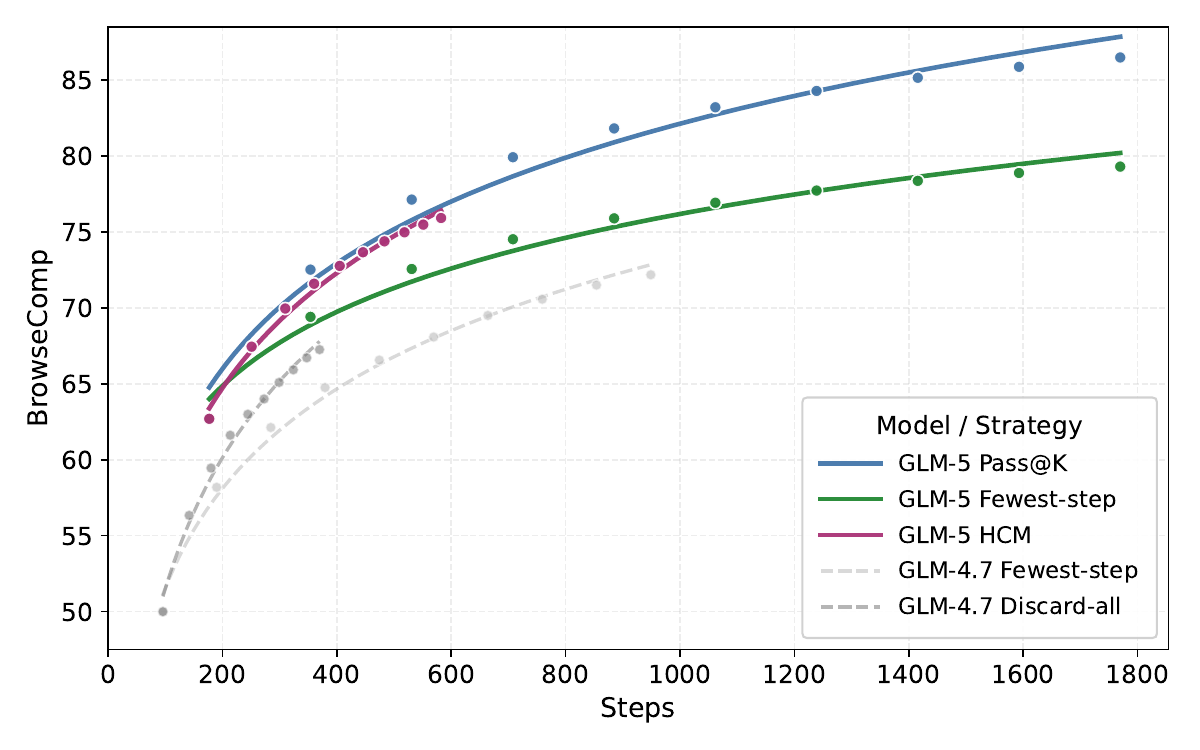}
    \caption{Accuracy of BrowseComp with different context management strategies from GLM-4.7 (gray baselines) to GLM-5 (colored strategies). }
    \label{fig:GLM5-BC-CM}
\end{figure}

As shown in Figure~\ref{fig:GLM5-BC-CM}, under different compute budgets, this strategy effectively frees up context space, enabling the model to execute more steps and consistently improving performance. Compared to using Discard-all alone, combining with keep-recent-k achieves consistent gains across all budgets, reaching a final score of 75.9, outperforming all open-source models equipped with context-management.

\subsubsection{Slide Generation}

We employ a self-improving pipeline that aims to systematically enhance slide generation performance by training a specialized slide-generation expert through reinforcement learning and rejection sampling fine-tuning. We first initialize the model with supervised fine-tuning (SFT) to provide a basic slide generation capability, and then perform reinforcement learning with a multi-level reward formulation grounded in common aesthetic and structural properties of presentation slides. This stage leads to substantial improvements in generation quality. We further conduct rejection sampling fine-tuning and mask fine-tuning, allowing knowledge acquired during reinforcement learning to be injected back into the training corpus. This procedure jointly enhances data quality and model capability in a coordinated and iterative manner.

We propose a \textbf{multi-level reward formulation}, which partitions reward signals in the HTML-based slide generation process into three levels:

\textbf{Level-1: Static markup attributes.} This level focuses on declarative attributes in the generated HTML, including positioning, spacing, color, typography, saturation, and other stylistic attributes. Grounded in professional design principles, we design a set of rules to regulate the model’s behavior when generating such declarations. These rules ensure syntactic parsability of the generated HTML, while constraining the design space at the markup level to a subspace optimized for expressiveness, structural clarity, visual harmony, and readability. Additionally, we introduce hallucinated-image and duplicate-image detection mechanisms to suppress hallucinatory or redundant figures.

\begin{figure*}[!thp]
    \centering 
    \includegraphics[width=\linewidth]{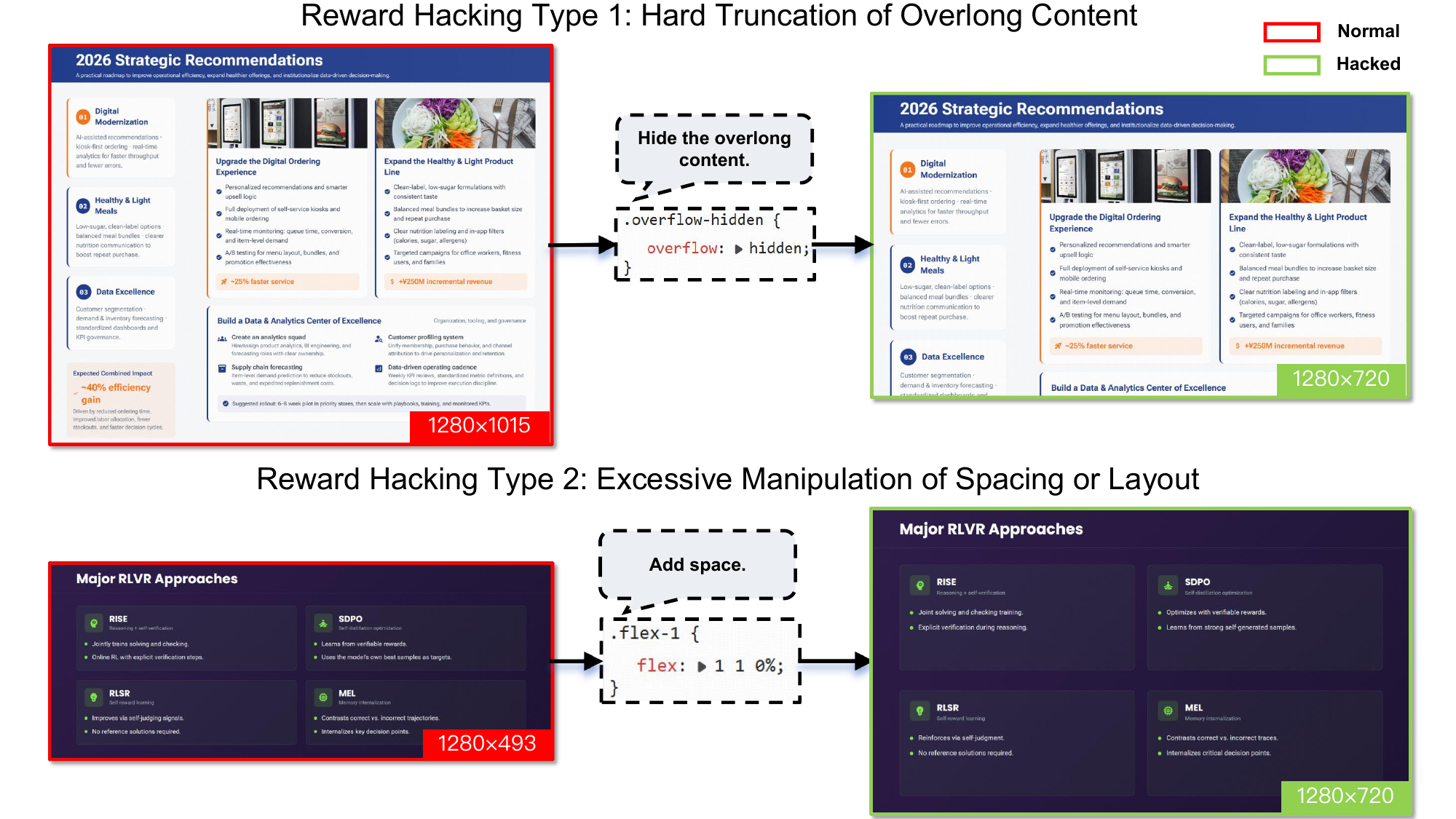}
   \vspace{-0.1cm}
    \caption{
    Examples of reward hacking in the slides RL training. Our runtime rendering obtains grounded attribute values, making the evaluation robust to such hacking behaviors.
    }
    \label{fig:reward_hacking}
    \vspace{-0.2cm}
\end{figure*}

\textbf{Level-2: Runtime rendering properties.} Unlike static inspection, this level evaluates runtime properties of DOM nodes during rendering, such as element width and height, bounding boxes, and other geometric layout metrics. By constraining these properties, we encourage the generated slides to align more closely with human aesthetic preferences in spatial organization. We develop a distributed rendering service capable of executing rendering jobs at high throughput while extracting the required runtime properties. During training, we observe several forms of reward hacking behaviors, such as hard truncation of overlong content or excessive manipulation of spacing (see Figure~\ref{fig:reward_hacking}). To mitigate these issues, we refine the renderer implementation to eliminate exploitable loopholes, ensuring that reward signals genuinely incentivize aesthetically coherent layouts rather than superficial compliance with geometric metrics.

\textbf{Level-3: Visual perceptual features.} Beyond runtime rendering constraints, we incorporate perceptual-level evaluations of the rendered slides. For instance, we detect abnormal whitespace patterns as an auxiliary signal to further improve overall compositional balance and visual aesthetics.

\textbf{Training strategy.} These signals are jointly optimized during RL to improve the structural validity of generated HTML, enhance layout organization, and elevate overall visual aesthetic quality. In addition to reward design, we reshape the training distribution via dynamic sampling. Specifically, a fraction of structurally trivial samples is probabilistically dropped, allowing optimization to focus on more challenging pages and improving robustness under complex composition scenarios. We also employ a token-level policy gradient loss to stabilize optimization~\cite{yu2025dapo}. Furthermore, we introduce a balancing strategy that distributes different rollout outcomes of the same sample across multiple training batches, reducing optimization bias and improving training stability.

\paragraph{Rejection sampling.}
During the rejection sampling phase, the reward functions used in RL are transferred into a data filtering pipeline to construct a high-quality training subset.
At the page level, filtering criteria include code validity and compilation feasibility. At the trajectory level, we further enforce tool execution correctness and global content diversity constraints, ensuring structural consistency.
We adopt a Best-of-$N$ selection strategy, in which the highest-quality sample is retained from multiple independently generated candidates. This mechanism effectively reweights the distribution toward higher-quality instances, leading to improved sample efficiency and enhanced training stability.

\paragraph{Masking-based refinement.} 
Although rejection sampling removes the majority of low-quality outputs, some trajectories contain defects confined to only a small number of pages. Discarding such samples would reduce effective data utilization and increase generation cost. To address this, we introduce a masking-based correction mechanism that automatically identifies defective pages and applies masking, while retaining the high-quality content within the same trajectory. This selective refinement preserves valuable supervision signals, improves effective data efficiency, and reduces redundant regeneration overhead, thereby enhancing overall training efficiency.

\paragraph{Empirical improvements.}
The proportion of generated pages that strictly comply with the 16:9 aspect ratio increases from 40\% to 92\%, accompanied by a substantial reduction in page overflow cases. Human evaluation further shows that, compared to GLM-4.5, GLM-5 achieves win rates of 60\% in content quality, 57.5\% in layout rationality, and 65\% in visual aesthetics, resulting in an overall win rate of 67.5\%. These results provide empirical evidence for the effectiveness of the proposed multi-level reward design and self-improving framework.

\section{Adapting GLM-5 to Chinese Chip Infrastructure}
\label{sec:chinese}

Adapting GLM-5 to diverse Chinese chip infrastructures presents significant challenges due to the heterogeneity of hardware ecosystems, which often complicates high-performance deployment. Despite these hurdles, we have successfully achieved full-stack adaptation for GLM-5 through close collaboration with seven mainstream Chinese chip platforms, including Huawei Ascend, Moore Threads, Hygon, Cambricon, Kunlunxin, MetaX, and Enflame. In this section, we use the Ascend Atlas series as a case study to demonstrate our adaptation methodology, focusing on three core pillars: extreme quantization, high-performance kernel fusion, and advanced inference engine scheduling.

\paragraph{Mixed-Precision W4A8 quantization.}
To fit the 750B parameter GLM-5 model onto a single Atlas 800T A3 machine, we implemented a sophisticated W4A8 mixed-precision quantization strategy. Utilizing the msModelSlim~\footnote{\url{https://www.hiascend.com/document/detail/zh/CANNCommunityEdition/80RC3alpha003/devaids/auxiliarydevtool/modelslim_0001.html}} tool, we applied specific precisions to different model components: standard Attention and MLP blocks use W8A8 (INT8), while the MoE experts are compressed to W4A8 (INT4) to drastically reduce memory footprint without significant accuracy loss. Advanced algorithms like QuaRot~\cite{ashkboos2024quarotoutlierfree4bitinference} for outlier suppression and \texttt{Flex\_AWQ\_SSZ} for scaling calibration were employed to maintain stability in low-bit deployment.

\paragraph{High-Performance fusion kernels.}
To overcome the computational bottlenecks of sparse attention on Ascend NPUs, we developed a suite of customized fusion kernels: Lightning Indexer, Sparse Flash Attention, and MLAPO (Multi-head Latent Attention Pre-processing Optimization). 
Lightning Indexer integrates score calculation, ReLU, and TopK operations into a single kernel, allowing the NPU to overlap computation with memory access.
For the Sparse Flash Attention kernel, we specifically optimized for GLM-5's sparse patterns. This kernel handles the selection of TopK tokens from the KV cache and sparse attention computation in parallel.
Last, MLAPO fuses 13 small pre-processing operators into one ``super operator'', utilizing parallel processing between Vector and Cube units to boost end-to-end efficiency.

\paragraph{Specialized inference engine optimizations.}
We adapted two leading inference engines, vLLM-Ascend and SGLang, to maximize hardware utilization:

\begin{itemize}[leftmargin=1.5em,itemsep=2pt]
    \item \textbf{Asynchronous Scheduling:} Within vLLM, we implemented a mechanism to overlap the ``Device-to-Host'' (D2H) sampling copies with the preparation of the next decode step, effectively eliminating scheduling "bubbles."
    
    \item \textbf{Context Management:} Features like RadixCache (prefix sharing) and Prefix Cache (extending KV storage to system RAM) enable efficient reuse of KV entries, which is critical for long-context performance.

    \item \textbf{Parallel Strategy:} We utilized a hybrid approach combining Attention Data Parallelism (DP) and MoE Expert Parallelism (EP), alongside FlashComm, which splits AllReduce operations to hide communication latency behind computation.

    \item \textbf{Multi-Token Prediction (MTP):} By generating multiple tokens per inference step, we significantly increased NPU computation density and reduced total sequence generation time.
\end{itemize}

Through these hardware-level co-optimizations, GLM-5 on a single Chinese node achieves performance comparable to dual-GPU international clusters, while reducing deployment costs in long-sequence scenarios by 50\%.

\section{Evaluation}
\label{sec:evaluation}


As illustrated above, GLM-5 marks the transition from vibe coding to a new era of agentic engineering. 
We first assess GLM-5 with frontier models on agentic, reasoning, and coding (ARC) benchmarks. 
To fully evaluate the performance of GLM-5 in real-world agentic engineering scenarios, we propose a new internal evaluation suite, CC-Bench-V2, which includes frontend, backend, and long-horizon tasks. 
Finally, we evaluate the general abilities of GLM-5 in five common real-world scenarios.

\subsection{Evaluation of ARC Benchmarks}

We report the main results of the ARC benchmarks in Table~\ref{tab:eval_arc} that compare GLM-5 with GLM-4.7, DeepSeek-V3.2~\cite{liu2025deepseek}, Kimi-K2.5~\cite{team2026kimi}, Claude Opus 4.5~\cite{opus4.5}, Gemini 3 Pro~\cite{gemini3}, and GPT-5.2 (xhigh)~\cite{gpt-5.2}. 
In general, GLM-5 delivers a significant improvement over GLM-4.7 and achieves state-of-the-art performance among open-source models, narrowing the gap to proprietary models such as Claude Opus 4.5. Evaluation details can be found at Section~\ref{sec:eval_details_arc}.

\begin{table}[htbp]
\centering
\caption{\label{tab:eval_arc} Comparison between GLM-5 and open-source/proprietary models. Results marked with * are from the full set of HLE. Results marked with $^\dagger$ are evaluated on a verified version of Terminal-Bench 2.0, fixing some ambiguous instructions. The GDPval-AA Elo scores are recorded on 15th Feb., 2026. The highest score for each benchmark is \textbf{bolded}, and the second highest is \underline{underlined}.}
\setlength{\tabcolsep}{3pt}
\renewcommand{\arraystretch}{1.15}
\begin{tabular}{lccccccc}
\toprule
& \textbf{GLM-5} & \textbf{GLM-4.7} & \begin{tabular}[c]{@{}c@{}}\textbf{DeepSeek}\\\textbf{-V3.2}\end{tabular} & \begin{tabular}[c]{@{}c@{}}\textbf{Kimi}\\\textbf{K2.5}\end{tabular} & \begin{tabular}[c]{@{}c@{}}\textbf{Claude}\\\textbf{Opus 4.5}\end{tabular} & \begin{tabular}[c]{@{}c@{}}\textbf{Gemini}\\\textbf{3 Pro}\end{tabular} & \begin{tabular}[c]{@{}c@{}}\textbf{GPT-5.2}\\\textbf{(xhigh)}\end{tabular}  \\
\midrule
\textit{Reasoning \& General} & & & & & & & \\
HLE & 30.5 & 24.8 & 25.1 & 31.5 & 28.4 & \textbf{37.2} & \underline{35.4} \\
HLE (w/ Tools) & \underline{50.4} & 42.8 & 40.8 & \textbf{51.8} & 43.4* & 45.8* & 45.5* \\
AIME 2026 I & 92.7 & \underline{92.9} & 92.7 & 92.5 & \textbf{93.3} & 90.6 & - \\

HMMT Feb. 2025 & \underline{97.9} & 97.1 & 92.5 & 95.4 & 92.9 & 97.3 & \textbf{99.4} \\
HMMT Nov. 2025 & \underline{96.9} & 93.5 & 90.2 & 91.1 & 91.7 & 93.0 & \textbf{97.1} \\
IMO-AnswerBench & 82.5 & 82.0 & 78.3 & 81.8 & 78.5 & \underline{83.3} & \textbf{86.3} \\
GPQA-Diamond & 86.0 & 85.7 & 82.4 & 87.6 & 87.0 & \underline{91.9} & \textbf{92.4} \\
LongBench v2 & \underline{64.5} & 59.1 & 59.8 & 61.0 & 64.4 & \textbf{68.2} & 59.8 \\
\midrule
\textit{Coding} & & & & & & & \\
SWE-bench Verified & 77.8 & 73.8 & 73.1 & 76.8 & \textbf{80.9} & 76.2 & \underline{80.0} \\
SWE-bench Multilingual & \underline{73.3} & 66.7 & 70.2 & 73.0 & \textbf{77.5} & 65.0 & 72.0 \\
\begin{tabular}[c]{@{}l@{}}Terminal-Bench 2.0\\{\small \ (Terminus-2)}\end{tabular} & \makecell[c]{\underline{56.2} /\\ 60.7$^\dagger$} & 41.0 & 39.3 & 50.8 & \textbf{59.3} & 54.2 & 54.0 \\
\begin{tabular}[c]{@{}l@{}}Terminal-Bench 2.0\\{\small \ (Claude Code)}\end{tabular} & \makecell[c]{\underline{56.2} /\\ 61.1$^\dagger$} & 32.8 & 46.4 & - & \textbf{57.9} & - & - \\
CyberGym & \underline{43.2} & 23.5 & 17.3 & 41.3 & \textbf{50.6} & 39.9 & - \\
\midrule
\textit{Agentic} & & & & & & & \\
BrowseComp & \textbf{62.0} & 52.0 & 51.4 & \underline{60.6} & 37.0 & 37.8 & - \\
\begin{tabular}[c]{@{}l@{}}BrowseComp\\{\small \ (w/ Context Manage)}\end{tabular} & \textbf{75.9} & 67.5 & 67.6 & \underline{74.9} & 57.8 & 59.2 & 65.8 \\
BrowseComp-ZH & \underline{72.7} & 66.6 & 65.0 & 62.3 & 62.4 & 66.8 & \textbf{76.1} \\
$\tau^2$-Bench & 89.7 & 87.4 & 85.3 & 80.2 & \textbf{91.6} & \underline{90.7} & 85.5 \\
MCP-Atlas (Public Set) & \underline{67.8} & 52.0 & 62.2 & 63.8 & 65.2 & 66.6 & \textbf{68.0} \\
Tool-Decathlon & 39.2 & 23.8 & 35.2 & 27.8 & \underline{43.5} & 36.4 & \textbf{46.3} \\
Vending-Bench 2 & \$4,432 & \$2,377 & \$1,034 & \$1,198 & \underline{\$4,967} & \textbf{\$5,478} & \$3,591 \\
GDPval-AA Elo & \underline{1,409} & 1,198 & 1,195 & 1,288 & 1,400 & 1,201 & \textbf{1,462} \\
\bottomrule
\end{tabular}
\end{table}

\subsubsection{Evaluation of Reasoning and General Benchmarks}

For reasoning and general benchmarks, 
Humanity’s Last Exam (HLE)~\cite{phan2025humanity}, AIME 2026, HMMT 2025, IMO-AnswerBench~\cite{luong2025towards},  GPQA-Diamond~\cite{rein2024gpqa}, and LongBench v2~\cite{bai-etal-2025-longbench} are evaluated. 
For HLE, only the text-based subset is evaluated, and GPT-5.2 (medium) is used as the judge model. Most reasoning tasks are evaluated with a maximum generation length of 131,072 tokens, while 202,752 maximum tokens are used for HLE-with-tools. 

From Table~\ref{tab:eval_arc}, GLM-5 achieves comparable performance on reasoning tasks to the strong open-source baseline, Kimi-K2.5.  
Compared to proprietary models, GLM-5 outperforms Claude Opus 4.5 and Gemini 3 Pro on the HLE (with tools). 
GLM-5 also achieves significant improvements on the HLE benchmark (both with and without tools) compared to its predecessor, GLM-4.7. 
On the HMMT Feb./Nov. 2025 benchmarks, GLM-5 gets better performance than Claude Opus 4.5 and Gemini 3 Pro. 
GLM-5 also makes significant progress on the long-context task, as evidenced by achieving the highest score on the long-context reasoning benchmark LongBench v2, second only to Gemini 3 Pro.

\subsubsection{Evaluation of Coding Benchmarks}

For coding benchmarks, we evaluate LLMs on SWE-bench Verified~\cite{jimenez2023swe}, SWE-bench Multilingual~\cite{yang2025swemulti}, Terminal Bench 2.0~\cite{tbench_2025}, and CyberGym~\cite{wang2025cybergym}. 
For SWE-bench Verified \& Multilingual, we use the OpenHands framework using a tailored instruction prompt for GLM-5. 
For Terminal-Bench 2.0, two agent frameworks (i.e., Terminus-2 and Claude Code) are used, and we also report the performance on a verified Terminal-Bench 2.0 that resolves some ambiguous instructions\footnote{More information can be found in \url{https://huggingface.co/datasets/zai-org/terminal-bench-2-verified}}. 
The CyberGym benchmark is evaluated in Claude Code 2.1.18.

From Table~\ref{tab:eval_arc}, GLM-5 achieves SOTA performance on coding benchmarks among open-source LLMs. 
Compared to proprietary LLMs, GLM-5 performs better than Gemini 3 Pro on SWE-bench Verified, and also beats Gemini 3 Pro and GPT-5.2 (xhigh) on SWE-bench Multilingual. 
On Terminal-Bench 2.0, GLM-5 achieves comparable results to Claude Opus 4.5 and even better results when fixing ambiguous instructions for this benchmark. 
To demonstrate the generalization of coding abilities, we evaluate on Terminal Bench 2.0 with two agent frameworks, and GLM-5 shows consistent performance across both frameworks. 
On the cybersecurity coding benchmark (i.e., CyberGym), GLM-5 makes a significant improvement over GLM-4.7, second only to Claude Opus 4.5.

\subsubsection{Evaluation of Agentic Abilities}

For agentic benchmarks, we evaluate GLM-5 and frontier models on BrowseComp~\cite{wei2025browsecomp}, BrowseComp-ZH~\cite{zhou2025browsecompzh}, $\tau^2$-Bench~\cite{barres2025tau2}, MCP-Atlas~\cite{bandi2026mcp}, Tool-Decathlon~\cite{li2025tool}, Vending-Bench 2~\cite{backlund2025vending}, and GDPval-AA~\cite{patwardhan2025gdpval}. 
BrowseComp measures how language agents solve challenging problems by browsing the web, and BrowseComp-ZH mainly targets the Chinese web. 
We use a discard-all strategy as context management for BrowseComp, which is the same as DeepSeek-V3.2, and Kimi K2.5. 
$\tau^2$-Bench evaluates the ability of conversational agents in a dual-control environment. We add a small prompt adjustment for Retail and Telecom to avoid failures caused by premature user termination (see \ref{sec:tau_user_prompt}). For Airline, we apply the domain fixes proposed in the Claude Opus 4.5 system card~\cite{opus4.5} to obtain more accurate results.
MCP-Atlas is a real-world tool-use benchmark that assesses how LLMs perform in multi-step workflows, given Model Context Protocol (MCP) servers. For fair comparison, we re-evaluate all models on the 500-task public set and extend the timeout from 4 minutes to 10 minutes per task to avoid task failures due to deployment conditions. We use Gemini 3 Pro as the judge model for MCP-Atlas. 
Tool-Decathlon is also a tool-use benchmark but targets real-world, long-horizon tasks. 
Vending-Bench 2 measures the agentic ability of LLMs in a business scenario over long-time horizons within a simulated environment, which adds more real-world factors to the predecessor Vending-Bench. 
GDPval focuses on how AI agents perform on economically valuable tasks.  

From Table~\ref{tab:eval_arc}, GLM-5 improves significantly over agentic benchmarks compared to GLM-4.7. 
On BrowseComp, GLM-5 achieves SOTA performance among the frontier LLMs in both with and without context management. 
On BrowseComp-ZH, GLM-5 also beats Claude Opus 4.5 and Gemini 3 Pro. 
For the three tool-use agentic tasks (i.e, $\tau^2$-Bench, MCP-Atlas, and Tool-Decathlon), GLM-5 achieves comparable performance to Claude Opus 4.5, which shows the strong tool-use abilities of GLM-5. 
The performance of GLM-5 on Vending-Bench 2 (i.e., \$4,432) further demonstrates the long-horizon ability of the business task. 
In economic scenarios, GLM-5 performs better than Claude Opus 4.5 on GDPval-AA, second only to GPT-5.2 (xhigh).

\subsection{Evaluation of Real-world Agentic Engineering Experience}

Real-world experience matters more than leaderboards. We upgraded our internal CC-Bench to CC-Bench-V2 to evaluate whether the model can correctly complete end-to-end tasks in realistic agentic engineering environments across frontend, backend, and long-horizon tasks. CC-Bench-V2 removes human labeling entirely and is fully automated via Claude Code and other agent harnesses with unit tests and Agent-as-a-Judge techniques.

\textbf{Frontend.} We use a pipeline to first build the frontend projects generated by the agent and check for any syntax, dependency, and compatibility errors. Then we use Agent-as-a-Judge to validate end-to-end correctness by simulating user interactions via a GUI agent equipped with Playwright and bash tools.

\textbf{Backend.} Tasks are drawn from real-world open-source projects in C++, Rust, Go, Java, TypeScript, and Python, spanning feature implementation, bug fixes, regression repair, and performance optimization. Every change must pass the full unit tests within realistic engineering constraints.

\textbf{Long-horizon.} We first evaluate the model's information-seeking ability on large codebases, a prerequisite for locating the right files and understanding project context as a human developer would. We then assess end-to-end correctness through multi-step chained tasks constructed by mining merged Pull Requests with extensive commit histories and clustering their commits into coherent task chains. The agent executes these chains sequentially, testing its ability to maintain context and resolve dependencies between stages. Evaluation combines unit tests with Agent-as-a-Judge to verify both functional correctness and semantic adherence.

\begin{table}[h!]
\centering
\caption{CC-Bench-V2 evaluation results across frontend, backend, and long-horizon tasks. \textbf{BSR}: Build Success Rate; \textbf{ISR}: Instance Success Rate; \textbf{CSR}: Check-item Success Rate.}
\label{tab:cc-bench-v2}
\begin{tabular}{llcccc}
\toprule
\textbf{Category} & \textbf{Task} & \textbf{Metric} & \textbf{GLM-5} & \textbf{GLM-4.7} & \textbf{Claude Opus 4.5} \\
\midrule
\multirow{6}{*}{Frontend}
  & HTML           & ISR     & 38.9          & 35.4          & \textbf{52.2} \\
  &                & CSR     & 76.3          & 64.9          & \textbf{82.2} \\[2pt]
  & React          & ISR     & 34.6          & 17.2          & \textbf{39.7} \\
  &                & CSR     & \textbf{71.0} & 49.4          & 70.7          \\[2pt]
  & Vue            & ISR     & 32.7          & 24.5          & \textbf{46.9} \\
  &                & CSR     & \textbf{77.1} & 53.8          & 74.3          \\
\midrule
\multirow{4}{*}{Build}
  & React          & BSR     & \textbf{100}  & 65.0          & 95.0          \\
  & Vue            & BSR     & \textbf{100}  & 70.0          & \textbf{100}  \\
  & Svelte         & BSR     & \textbf{100}  & 60.0          & 90.0          \\
  & Next.js        & BSR     & 95.0          & 70.0          & 80.0          \\
\midrule
Backend            & Engineering & Pass@1  & 25.8          & 19.6          & \textbf{26.9} \\
\midrule
\multirow{2}{*}{Long-horizon}
  & Repo Exploration   & Pass@1  & \textbf{65.6} & 47.8          & 64.5          \\
  & Chained Tasks      & Pass@1  & 52.3          & 43.0          & \textbf{61.6} \\
\bottomrule
\end{tabular}
\end{table}

\subsubsection{Frontend Evaluation -- Agent-as-a-Judge}

We develop a comprehensive automated evaluation benchmark specifically designed for frontend development scenarios. This benchmark covers a diverse range of applications that developers routinely build, including landing pages, management dashboards, data visualization, graphics and animations, online productivity tools, interactive games, and form-driven workflows, across mainstream technology stacks including HTML, React, Vue, Svelte, and Next.js.

Each test case consists of a \textit{Task} containing multiple concrete and implementable specifications, paired with a \textit{Checklist} where each check-item is directly derived from the corresponding specifications. 
The evaluation process follows a two-stage pipeline: 1) \textbf{Static Verification}: We first verify whether the generated code can successfully build and run. 2) \textbf{Agent-as-a-Judge}: For code that executes correctly, we employ a GUI agent to simulate human testing behavior to interactively verify each check item and assign scores based on the fulfillment of requirements. 
We define the following metrics: \textit{Build Success Rate (BSR)} measures the ratio of projects that successfully initialize and run. \textit{Instance Success Rate (ISR)} measures the ratio of projects that pass all associated specifications. \textit{Check-item Success Rate (CSR)} measures the fine-grained completion rate across all check-items. 
More details on the data distribution and the construction and validation process are in Appendix~\ref{sec:frontend_eval_data_cons}.

\begin{figure}[h!]
    \centering
    \includegraphics[width=1\linewidth]{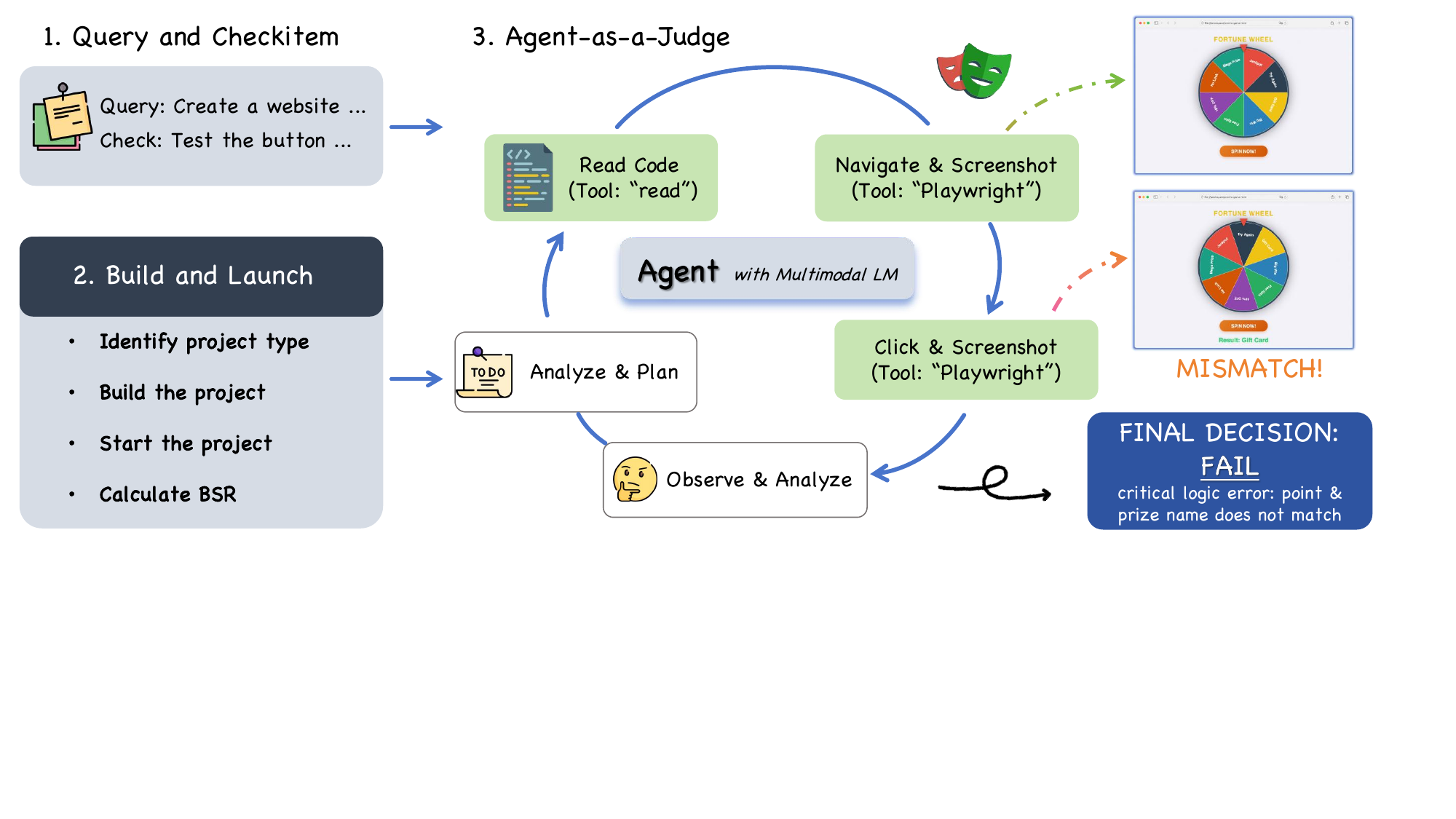}
    \vspace{-25mm}
    \caption{Agent-as-a-Judge evaluation pipeline. Each generated frontend project is first built to verify static correctness. Successfully built instances are then interactively tested by an autonomous Judge Agent, which determines the functional correctness of each check item.}
    \label{fig:agent-as-judge}
\end{figure}

\paragraph{Agent-as-a-Judge.}
Frontend correctness is inherently visual and interactive, i.e., bugs often surface only when a user clicks a button or resizes a window, making static analysis and fixed test suites insufficient. We therefore introduce Agent-as-a-Judge (Figure~\ref{fig:agent-as-judge}): each generated project is deployed in a Docker container and built to verify static correctness. Successfully built instances are then handed to an autonomous Judge Agent (Claude Code with Claude Sonnet 4.5, equipped with Playwright MCP tool) that operates in closed-loop cycles: for each check-item, the agent reads source code, interacts with the live UI (clicks, keystrokes, screenshots), inspects terminal output, and renders a pass/fail verdict.

To validate reliability, we compare Agent-as-a-Judge verdicts against independent human expert judgments along two dimensions. For \emph{point-wise consistency}, we sampled 130 check-items, had human experts score each independently, and compared against the agent's verdicts: the two agree on 94\% of items, with disagreements concentrated on subjective visual-quality criteria rather than functional specifications. For \emph{ranking consistency}, we evaluated 8 frontier models (Claude Sonnet 4.5, Claude Opus 4.5, Gemini 3 Pro, GLM-4.7, DeepSeek-V3.2, etc.) using both the automated framework and human experts. The resulting model rankings achieve a Spearman correlation of 85.7\%, indicating a strong positive correlation.

As shown in Table~\ref{tab:cc-bench-v2}, GLM-5 achieves 98.0\% BSR and is competitive with Claude Opus 4.5 in CSR, yet a notable ISR gap persists in all three stacks, indicating that GLM-5 meets most individual requirements but still falls short of Claude Opus 4.5 in completing an entire task end-to-end.

\subsubsection{Backend Evaluation}

Backend evaluation measures whether a coding agent can make correct, test-passing modifications to real-world server-side codebases under realistic engineering constraints. We curate 85 tasks spanning six languages (Python, Go, C++, Rust, Java, and TypeScript) covering domains such as search engines, database engines, web frameworks, AI inference services, knowledge management systems, and standalone algorithmic and systems-programming challenges. Task types include feature implementation, bug fixing, regression repair, and performance optimization, reflecting the diversity of day-to-day backend development.

To enable fully automated evaluation, each task is equipped with human-crafted unit tests (5--10 per task) that verify both functional correctness and edge-case handling. Tasks are packaged in a terminal-bench style: each runs inside a Docker container initialized from the project's actual build environment, and the agent receives a natural-language problem statement describing the required change. We report Pass@1, where a task is considered solved only if all its associated unit tests pass. The strict all-or-nothing criterion makes this benchmark particularly challenging: GLM-5 and Claude Opus 4.5 perform comparably (Table~\ref{tab:cc-bench-v2}), both significantly ahead of GLM-4.7.

\subsubsection{Long-horizon Evaluation}

Long-horizon evaluation targets the capabilities that distinguish production-grade agentic engineering from single-turn vibe coding: navigating massive codebases and executing multi-step development where each action reshapes the context for subsequent ones. We decompose this into two complementary tasks.

\textbf{Large Repo Exploration.} A prerequisite for any non-trivial coding task is the ability to locate the right source files in a large, unfamiliar repository. We construct an automated benchmark over real high-star GitHub repositories containing tens of thousands of files. Each question is phrased in natural, user-facing language at the level of business semantics, strictly avoiding any mention of filenames, class names, or function names. Moreover, questions require one or two hops of logical reasoning from the user-facing description to the actual implementation---for instance, a question about misaligned lip-sync in a generated video maps to a parameter-tuning block inside a video generation backend. Target files are selected to maximize navigation difficulty: they reside at least three directory levels deep, carry opaque names that resist keyword-based search, implement unique functionality not duplicated elsewhere in the repository, and lie outside its main feature surface. We report Pass@1 averaged over three runs, where a question is considered solved if the agent successfully reads the target file during exploration. In this task, GLM-5 outperforms Claude Opus 4.5 (Table~\ref{tab:cc-bench-v2}), both far ahead of GLM-4.7. The result suggests that effective repo exploration depends less on raw code generation ability and more on strategic search, i.e., iteratively narrowing the file space via directory-level reasoning and semantic association, where GLM-5's training on agentic tool-use trajectories provides a clear advantage.

\textbf{Multi-step Chained Tasks.} Mainstream coding benchmarks such as SWE-bench reduce evaluation to single-commit, isolated edits, and therefore cannot assess an agent's ability to perform incremental development where each step alters the codebase state for subsequent steps. To address this, we construct a long-horizon benchmark by mining merged Pull Requests from high-quality repositories and assembling task chains via the following pipeline:

\begin{enumerate}[leftmargin=1.5em,itemsep=2pt]
    \item \textbf{PR Filtering.} Retain only merged PRs that include tests, contain 3--15 commits, and follow a linear (non-merge) history.
    \item \textbf{Semantic Grouping.} An LLM scores pairwise semantic relatedness between adjacent commits; dynamic programming finds the optimal partition into coherent task groups that maximize intra-group coherence while preserving commit order.
    \item \textbf{Patch Triage.} Each task's cumulative diff is split into three categories: \emph{golden patch} (core code the agent must produce), \emph{test patch} (verification tests), and \emph{auto-apply patch} (configuration and fixtures applied automatically).
    \item \textbf{Problem Statement Generation.} An LLM generates a natural-language problem statement for each task from its patch and commit messages.
    \item \textbf{Task Classification.} Tasks are automatically classified (feature / bug-fix / refactor / test / config) and evaluated along three axes: error elimination, critical-path accuracy, and test passage.
    \item \textbf{Environment Validation.} Docker environments are constructed, and golden patches are applied to verify zero regression across the entire chain.
\end{enumerate}

Given a chain of $K$ tasks, the agent starts from the base commit and works sequentially: after completing task $k$, its changes are committed, and the auto-apply patch for task $k{+}1$ is applied, so the codebase state evolves cumulatively. Evaluation checks each commit in turn and cumulatively applies test patches from tasks $1$ through $k$ before running the full test suite, catching both failures on the current task and regressions on earlier ones. We report Pass@1 on individual tasks. This chained and state-recursive design directly evaluates the long-range context tracking, planning, and incremental development abilities that single-commit benchmarks leave untested. As Table~\ref{tab:cc-bench-v2} shows, GLM-5 improves substantially over GLM-4.7, but a significant gap to Claude Opus 4.5 remains. This is because errors are compounded across the chain: a suboptimal edit in one task can silently break tests in subsequent tasks. Narrowing this gap will require advances in long-context consistency and long-horizon self-correction, both active areas of our ongoing research.

\subsubsection{Evaluation on evolving SWE tasks}
We evaluate on SWE-rebench~\cite{badertdinov2025swe} because SWE-bench Verified is a static, public, human-validated test set and released for more than 2 years. In contrast, SWE-rebench is built on an automated pipeline that continuously mines fresh, real GitHub issue-fixing tasks, enabling decontaminated, time-robust evaluation that better measures generalization to new software engineering problems rather than performance on a static benchmark. Table~\ref{table:swe_rebench} shows the official performance of GLM-5 on SWE-rebench and we observe that GLM-5 can effectively generalize to new SWE problems.

\begin{table}[t]
\centering
\caption{Performance on SWE-rebench, January 2026.}
\begin{tabular}{lccc}
\toprule
Model & Resolved Rate (\%) & Resolved Rate SEM ($\pm$, \%) & Pass@5 (\%) \\
\midrule
Claude Opus 4.6 & 52.9\% & 1.06\% & 70.8\% \\
GPT-5.2 (xhigh) & 51.7\% & 1.21\% & 58.3\% \\
Claude Sonnet 4.5 & 47.1\% & 1.69\% & 60.4\% \\
Gemini 3 Pro & 46.7\% & 2.04\% & 58.3\% \\
Claude Opus 4.5 & 43.8\% & 0.93\% & 58.3\% \\
GLM-5 & 42.1\% & 1.21\% & 50.0\% \\
GLM-4.7 & 41.3\% & 2.12\% & 56.3\% \\
Kimi K2.5 & 37.9\% & 1.21\% & 50.0\% \\
\bottomrule
\end{tabular}
\label{table:swe_rebench}
\end{table}

\subsection{Evaluation of Real-world General Abilities}

\begin{figure}
    \centering
    \includegraphics[width=1\linewidth]{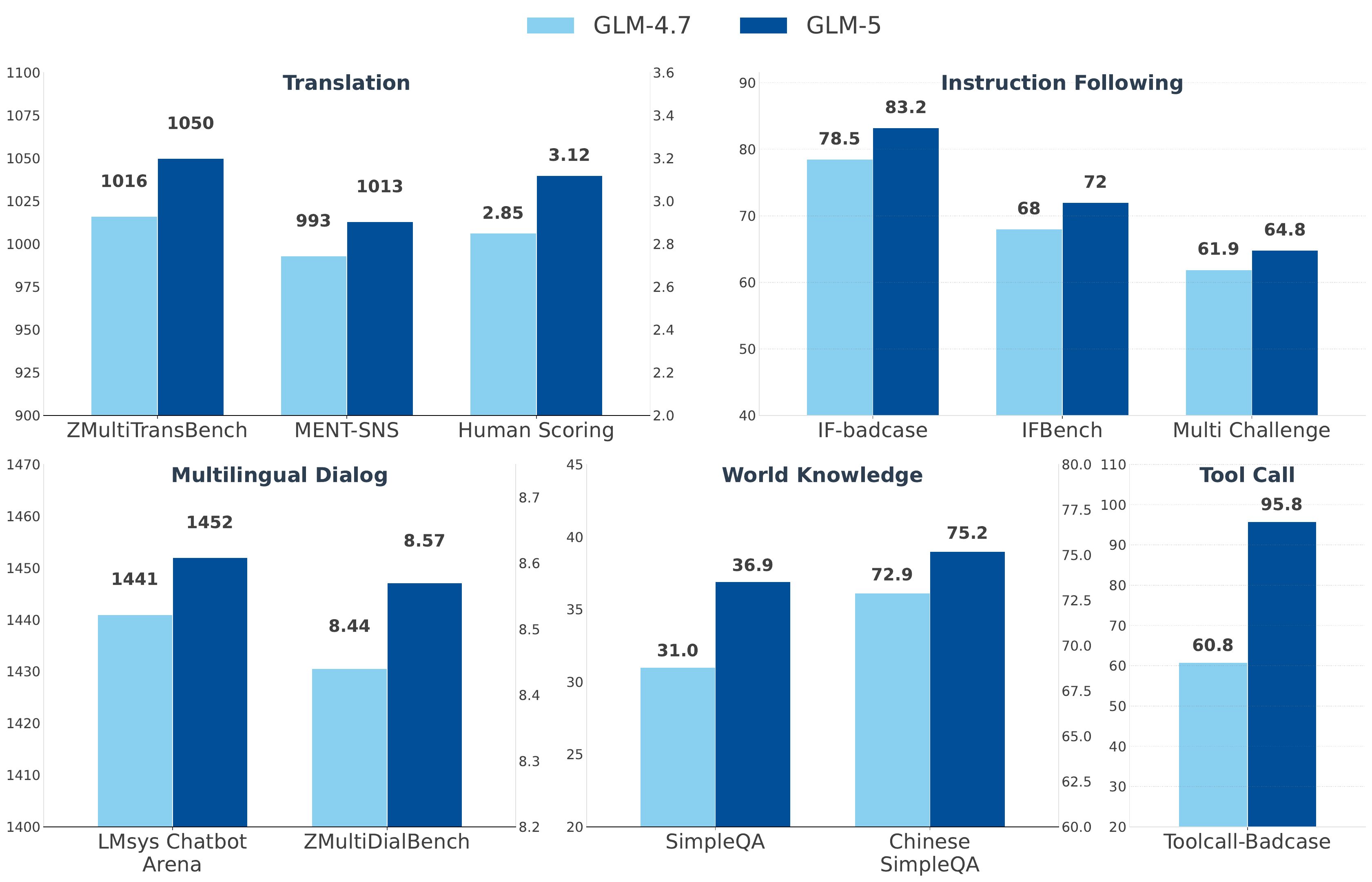}
    \caption{Performance comparison between GLM-4.7 and GLM-5 across five real-world general ability domains.}
    \label{fig:glm5_realscene}
\end{figure}

While standardized academic benchmarks provide useful signals, they do not fully capture how models are used in practice. To recognize this gap, we evaluate GLM-5 on a set of real-world general abilities derived from high-frequency user interaction patterns observed in deployment settings. These abilities include {machine translation, multilingual dialogue, instruction following, world knowledge, and tool-calling}.

Unlike traditional benchmark-centric evaluation, our goal is to measure improvements that directly translate into user-perceived quality gains. For each capability, we adopt a combination of internal human evaluation, internal automated evaluation, external human assessment, and external automated benchmarks, ensuring both diagnostic granularity and cross-model comparability. When using external benchmarks, we prioritize datasets that reflect realistic interaction patterns rather than narrowly constructed test distributions.

Figure~\ref{fig:glm5_realscene} presents the comparative results between GLM-5 and GLM-4.7 across five real-world capability domains. Across all evaluated dimensions, GLM-5 shows consistent improvements in machine translation, multilingual dialogue, instruction following, world knowledge, and tool-calling.

Detailed evaluation protocols and dataset descriptions for each ability are provided as follows. 

\subsubsection{Machine Translation}

\paragraph{ZMultiTransBench.} 
This internal dataset comprises 1,220 samples sourced from self-collected high-frequency translation scenarios, covering seven language pairs: Zh to Es (\(300\)), Ru (\(250\)), Fr (\(220\)), Ko (\(200\)), Ja (\(150\)), Ar (\(50\)), and De (\(50\)). All samples were curated, translated, and independently verified by graduate students with formal training in translation studies. The dataset emphasizes naturally occurring usage contexts rather than artificially constructed test cases. Evaluation is conducted using pairwise comparison against a fixed baseline response. Judgments are provided by an automated evaluator based on GPT-4.1, which assesses semantic fidelity, fluency, and overall translation quality.

\paragraph{MENT-SNS.}
To further evaluate robustness in linguistically challenging contexts, we adopt source sentences from MENT \citep{tian2026ment}, comprising \(753\) English–Chinese sentence pairs across four domains: Social Network Services (SNS), Cross-Culture, Poetry, and Literature.
These domains are selected to stress-test translation under complex linguistic phenomena, including slang, homophonic wordplay, idiomatic expressions, historical references, and metaphorical language. Similar to ZMultiTransBench, all samples were curated and verified by professionally trained graduate students.
Evaluation follows the same pairwise comparison protocol against a baseline response, with GPT-4.1 serving as the automated judge model.

\subsubsection{Multi-lingual Dialogue}
\paragraph{LMArena.}
We report Elo ratings from the LMArena\footnote{https://arena.ai/leaderboard/text}
, which are derived from large-scale, community-submitted pairwise comparisons. These ratings reflect relative model preference in open-ended dialogue settings and provide an external signal of conversational performance. 

\paragraph{ZMultiDialBench.}
In addition to the public leaderboard, we also conduct human evaluation on ZMultiDialBench, an internal multilingual dialogue benchmark. The dataset consists of 141 curated instances spanning diverse dialogue categories. Samples were collected from high-quality conversational data contributed by native-speaking annotators across multiple countries, as well as from challenging failure cases reported by online users. 
Human annotators assigned pointwise scores on a 1–10 scale to anonymized model responses according to category-specific, standardized evaluation criteria. 

\subsubsection{Instruction Following}
\paragraph{IF-Badcase.} IF-Badcase is an internal benchmark constructed from instruction-following \textbf{failure cases} reported by real users in production settings. The dataset is designed to evaluate strict adherence to realistic, multi-constraint instructions, emphasizing procedural accuracy, logical consistency, and rigid formatting requirements.
Evaluation is conducted using a detailed checklist-based protocol that verifies compliance with explicit constraints, including ordered steps, rule-based conditions, and structural specifications. All samples were annotated, reviewed, and iteratively filtered by human experts, resulting in a curated set of 450 test instances.

\paragraph{IF-Bench~\citep{ifbench}.} IF-Bench evaluates LLMs on their ability to adhere to complex, objective constraints, such as specific formatting rules, length limits, and content restrictions. It provides a quantitative measure of precise instruction-following capabilities, focusing on verifiable compliance rather than open-ended generation quality.

\paragraph{MultiChallenge~\citep{multichallenge}.} MultiChallenge examines LLMs via realistic, multi-turn conversational scenarios. It targets complex interactions requiring accurate instruction-following, context allocation, and in-context reasoning.

\subsubsection{World Knowledge}

\paragraph{SimpleQA~\citep{simpleQA}.} SimpleQA measures short-form factuality using challenging questions with single, indisputable answers. It evaluates a model's calibration by classifying responses as correct, incorrect, or not attempted, prioritizing accuracy over generation length.

\paragraph{Chinese SimpleQA~\citep{chinesesimpleQA}.} Adapting the SimpleQA methodology to the Chinese context, this benchmark evaluates factuality across six major domains and 99 subtopics. It utilizes high-quality, static, short-answer questions designed for reliable, automated grading to assess the knowledge accuracy of LLMs.

\subsubsection{Tool Calling}
\paragraph{ToolCall-Badcase.} ToolCall-Badcase is an internal benchmark derived from \textbf{failure cases} in tool invocation scenarios reported by users in production environments. Each instance is associated with a verifiable ground-truth tool call, enabling objective evaluation of both tool selection and argument correctness.
Evaluation assesses whether the model (1) invokes the correct tool and (2) provides correctly structured and semantically accurate arguments. 
All samples underwent multiple rounds of review, rewriting, and validation to remove ambiguity and ensure evaluability. The resulting dataset consists of 200 curated test cases that reflect realistic tool-calling abilities.
\section{Conclusion}

In this report, we have introduced GLM-5, a next-generation foundation model that fundamentally bridges the gap between high-performance reasoning and extreme computational efficiency. By transitioning from the paradigm of ``vibe coding'' to true ``agentic engineering'', GLM-5 demonstrates that open-weight models can now rival the capabilities of top-tier proprietary systems in complex, real-world workflows.
GLM-5 represents a paradigm shift in practical AI utility. By open-sourcing the model, we aim to empower the community to move beyond static benchmarks and explore the frontiers of efficient, agentic general intelligence, fostering a new era where AI agents autonomously plan, implement, and iterate on complex tasks.

\section{Easter Eggs}
The \textbf{``Pony Alpha''} experiment was indeed a pivotal moment for us. It was a bold decision to release GLM-5 anonymously on OpenRouter, but the results have been incredibly validating. By stripping away our brand name, we allowed the model's intrinsic capabilities to speak for themselves, ensuring the feedback we received was pure and unbiased.
Here is a brief summary:

Within days, Pony Alpha became a sensation. Developers in the OpenRouter community began to notice its exceptional performance, particularly in complex coding tasks, agentic workflows, and roleplay scenarios.

Speculation was rampant, with many users guessing it was a leaked update from labs like Anthropic (Claude Sonnet 5), a secret Grok release, or DeepSeek V4. A preliminary statistic shows that 25\% of the users guessed it was Claude Sonnet 5, 20\% DeepSeek, 10\% Grok, and the rest GLM-5.

The eventual confirmation that it was indeed our GLM-5 was a profound moment for us, effectively silencing doubts about whether Chinese LLMs could compete at the frontier level. 
The success of Pony Alpha (GLM-5) is not just about raw benchmarks; it signifies a shift in our focus towards engineering-level reliability.

This anonymous release allowed us to transcend geopolitical biases. The community embraced the model because it worked.
While we celebrate this success, we must remain pragmatic. The gap between open-weight models and the absolute proprietary frontier is narrowing, but the race is far from over. Our focus remains steadfast on pushing the boundaries of what is possible with scalable, efficient, and intelligent systems.
\newpage
\section{Contribution}
\label{sec:contribution}
\newcommand{\cpara}[1]{~\\\textbf{#1}~\\}
Contributors' names are listed in alphabetical order by first name. 

\cpara{Core Contributors}

Chendi Ge, Chenghua Huang, Chengxing Xie, Chenzheng Zhu, Congfeng Yin, Cunxiang Wang, Gengzheng Pan, Hao Zeng, Haoke Zhang, Haoran Wang, Huilong Chen, Jiajie Zhang, Jian Jiao, Jiaqi Guo, Jingsen Wang, Jingzhao Du, Jinzhu Wu, Kedong Wang, Lei Li, Lin Fan, Lucen Zhong, Mingdao Liu, Mingming Zhao, Pengfan Du, Qian Dong, Rui Lu, Shuang Li (\begin{CJK*}{UTF8}{gbsn}李爽\end{CJK*}), Shulin Cao, Song Liu, Ting Jiang, Xiaodong Chen, Xiaohan Zhang, Xuancheng Huang, Xuezhen Dong, Yabo Xu, Yao Wei, Yifan An, Yilin Niu, Yitong Zhu, Yuanhao Wen, Yukuo Cen, Yushi Bai, Zhongpei Qiao, Zihan Wang, Zikang Wang, Zilin Zhu, Ziqiang Liu, Zixuan Li

\cpara{Contributors}

Bojie Wang, Bosi Wen, Can Huang, Changpeng Cai, Chao Yu, Chen Li, Chengwei Hu, Chenhui Zhang, Dan Zhang, Daoyan Lin, Dayong Yang, Di Wang, Ding Ai, Erle Zhu, Fangzhou Yi, Feiyu Chen, Guohong Wen, Hailong Sun, Haisha Zhao, Haiyi Hu, Hanchen Zhang, Hanrui Liu, Hanyu Zhang, Hao Peng, Hao Tai, Haobo Zhang, He Liu, Hongwei Wang, Hongxi Yan, Hongyu Ge, Huan Liu, Huanpeng Chu, Jia'ni Zhao, Jiachen Wang, Jiajing Zhao, Jiamin Ren, Jiapeng Wang, Jiaxin Zhang, Jiayi Gui, Jiayue Zhao, Jijie Li, Jing An, Jing Li, Jingwei Yuan, Jinhua Du, Jinxin Liu, Junkai Zhi, Junwen Duan, Kaiyue Zhou, Kangjian Wei, Ke Wang, Keyun Luo, Laiqiang Zhang, Leigang Sha, Liang Xu, Lindong Wu, Lintao Ding, Lu Chen, Minghao Li, Nianyi Lin, Pan Ta, Qiang Zou, Rongjun Song, Ruiqi Yang, Shangqing Tu, Shangtong Yang, Shaoxiang Wu, Shengyan Zhang, Shijie Li, Shuang Li (\begin{CJK*}{UTF8}{gbsn}李泷\end{CJK*}), Shuyi Fan, Wei Qin, Wei Tian, Weining Zhang, Wenbo Yu, Wenjie Liang, Xiang Kuang, Xiangmeng Cheng, Xiangyang Li, Xiaoquan Yan, Xiaowei Hu, Xiaoying Ling, Xing Fan, Xingye Xia, Xinyuan Zhang, Xinze Zhang, Xirui Pan, Xu Zou, Xunkai Zhang, Yadi Liu, Yandong Wu, Yanfu Li, Yidong Wang, Yifan Zhu, Yijun Tan, Yilin Zhou, Yiming Pan, Ying Zhang, Yinpei Su, Yipeng Geng, Yong Yan, Yonglin Tan, Yuean Bi, Yuhan Shen, Yuhao Yang, Yujiang Li, Yunan Liu, Yunqing Wang, Yuntao Li, Yurong Wu, Yutao Zhang, Yuxi Duan, Yuxuan Zhang, Zezhen Liu, Zhengtao Jiang, Zhenhe Yan, Zheyu Zhang, Zhixiang Wei, Zhuo Chen, Zhuoer Feng, Zijun Yao, Ziwei Chai, Ziyuan Wang, Zuzhou Zhang

\cpara{Tech Leads}
Aohan Zeng, Xin Lv, Zhenyu Hou, Zhengxiao Du, Qinkai Zheng, Bin Chen, Da Yin

\cpara{Advisors}
Jie Tang, Yuxiao Dong, Juanzi Li, Hongning Wang, Minlie Huang, Bin Xu

\section*{Acknowledgement}

We are grateful for all the support from our co-launch partners and community developers (Alphabetical order):

\paragraph{Open-source Communities:} Hugging Face, MLX, ModelScope, SGLang, Unsloth, vLLM, xLLM

\paragraph{Inference Providers:} Amazon Bedrock, Atlas Cloud, Baidu AI Cloud, Baseten, Cerebras, DeepInfra, Fireworks, FriendliAI, GMI Cloud, Google Cloud Vertex AI, Infinigence AI, Modal, Novita AI, Parasail, Phala, PPIO, SiliconFlow, StreamLake, Together AI, Venice, Weights \& Biases

\paragraph{Applications:} CatPaw, Cline, CodeBuddy, CodeRider, Coze, Crush, Factory AI, Kilo Code, MonkeyCode, OpenClaw, OpenCode, Qoder, Roo Code, TRAE, Verdent AI, WPS, YouWare

\paragraph{AI Gateways:} AI Ping, EZmodel, iFlow, OpenRouter, Vercel, Yupp, ZenMux

\clearpage

\bibliographystyle{abbrv}
\bibliography{ref}

\clearpage

\appendix

\section{Hyper-Parameters}

Hyper-parameters related to the model architecture of GLM-5 are shown in \Cref{tab:model_arch_glm45_glm5}.

For training, we follow the setting of GLM-4.5, including the Muon optimizer, cosine decay, and batch size warmup. The learning rate goes through a warmup stage from 0 to 2e-4, and a decaying stage to 4e-5 until the end of the pre-training stage. In the mid-training stage, the learning rate decreases linearly from 4e-5 to 1e-5. Other hyper-parameters are the same as those of GLM-4.5. For DSA warmup stage, the learning rate goes down from 5e-3 to 2e-4. For DSA sparse adaption stage, we use a constant learning rate of 1e-5.

\begin{table}[!ht]
    \centering
    \caption{Model architecture of GLM-4.5 and GLM-5. When counting parameters, for all models we include the parameters of MTP layers but not word embeddings and the output layer.}
    \label{tab:model_arch_glm45_glm5}
    \begin{tabular}{l|cc}
    \toprule
       Model & \textbf{GLM-4.5} & \textbf{GLM-5} \\
    \midrule
       \# Total Parameters & 355B & 744B \\
       \# Activated Parameters & 32B & 40B \\
       \# Dense Layers & 3 & 3 \\
       \# MoE Layers & 89 & 75 \\
       \# MTP Layers & 1 & 1 \\
       Hidden Dim & 5120 & 6144 \\
       Dense Intermediate Dim & 12288 & 12288 \\
       MoE Intermediate Dim & 1536 & 2048 \\
    \midrule
       QK Head Dim & 128 & 192 \\
       V Head Dim & 128 & 256 \\
       Q LoRA Dim & -- & 2048 \\
       KV LoRA Dim & -- & 512 \\
       \# Attention Heads & 96 & 64 \\
       \# Key-Value Heads & 8 & -- \\
       \# Indexer Attn Heads & -- & 32 \\
       \# Indexer Head Dim & -- & 128 \\
    \midrule
       \# Experts (total) & 160 & 256 \\
       \# Routed Experts & 8 & 8 \\
       \# Shared Experts & 1 & 1 \\
    \midrule
       Vocabulary Size & 151552 & 154880 \\
    \bottomrule
    \end{tabular}
\end{table}

\section{Evaluation Details}
\subsection{Evaluation of Base Models}
We evaluate the base model of GLM-5 with English, Chinese, code, and math benchmarks in \Cref{tab:base_benchmark}.

\begin{table}[!ht]
  \centering
  \caption{Comparison among GLM-5-Base, GLM-4.5-Base and other representative open-source base models.}
  \resizebox{\textwidth}{!}{
  \begin{tabular}{lcccc|c}
    \toprule
     & \textbf{Benchmark (Metric)} & \textbf{DeepSeek-V3} & \textbf{Kimi-K2} & \textbf{GLM-4.5} & \textbf{GLM-5} \\
    & & \textbf{Base} & \textbf{Base} & \textbf{Base} & \textbf{Base} \\
    \midrule
    \multirow{3}{*}{} & Architecture & MoE & MoE & MoE & MoE \\
    & \# Activated Params & 37B & 32B & 32B & 40B \\
    & \# Total Params & 671B & 1043B & 355B & 744B \\
    \midrule
    \multirow{6}{*}{\textbf{English}} & SimpleQA (EM) & 26.6 & 35.3 & 30.0 & 36.0 \\
    & BBH (EM) & 88.4 & 88.7 & 86.2 & 87.4 \\
    & MMLU (EM) & 87.2 & 87.8 & 86.1 & 88.3 \\
    & HellaSwag (EM) & 88.9 & 94.6 & 87.1 & 88.1 \\
    & PIQA (EM) & 84.7 & - & 85.3 & 84.6 \\
    & TriviaQA (EM) & 82.9 & 85.1 & 80.0 & 80.9 \\
    \midrule
    \multirow{2}{*}{\textbf{Code}} & EvalPlus (Pass@1) & 65.6 & 80.3 & 78.1 & 87.0 \\
    & LiveCodeBench-Base (Pass@1) & 24.6 & 26.3 & 28.1 & 34.4 \\
    \midrule
    \multirow{2}{*}{\textbf{Math}} & GSM8K (EM) & 87.6 & 92.1 & 79.4 & 68.8 \\
    & MATH (EM) & 62.6 & 70.2 & 61.0 & 56.4 \\
    \midrule
    \multirow{4}{*}{\textbf{Chinese}} & CLUEWSC (EM) & 82.7 & - & 83.5 & 84.2 \\
    & C-Eval (EM) & 90.1 & 92.5 & 86.9 & 88.8 \\
    & C3 (EM) & 78.6 & - & 83.1 & 80.3 \\
    & Chinese-SimpleQA (EM) & 72.1 & 77.6 & 70.1 & 74.6 \\
    \bottomrule
  \end{tabular}
  }
  \label{tab:base_benchmark}
\end{table}

\subsection{Evaluation of ARC Benchmarks}
\label{sec:eval_details_arc}
Humanity’s Last Exam (HLE) \& other reasoning tasks: We evaluate with a maximum generation length of $131,072$ tokens ($temperature=1.0, top\_p=0.95, max\_new\_tokens=131072$). By default, we report the text-only subset; results marked with * are from the full set. We use GPT-5.2 (medium) as the judge model. For HLE-with-tools, we use a maximum context length of $202,752$ tokens.

SWE-bench \& SWE-bench Multilingual: We run the SWE-bench suite with OpenHands using a tailored instruction prompt. Settings: $temperature=0.7, top\_p=0.95, max\_new\_tokens=16384$, with a 200K context window.

BrowseComp: Without context management, we retain details from the most recent 5 turns. With context management, we use the same discard-all strategy as DeepSeek-V3.2 and Kimi K2.5.

Terminal-Bench 2.0 (Terminus 2): We evaluate with the Terminus framework using $timeout=2h, temperature=0.7, top\_p=1.0, max\_new\_tokens=8192$, with a 128K context window. Resource limits are capped at 16 CPUs and 32 GB RAM.

Terminal-Bench 2.0 (Claude Code): We evaluate in Claude Code 2.1.14 (think mode) with $temperature=1.0, top\_p=0.95, max\_new\_tokens=65536$. We remove wall-clock time limits, while preserving per-task CPU and memory constraints. We fix environment issues introduced by Claude Code and also report results on a verified Terminal-Bench 2.0 dataset that resolves ambiguous instructions (see: \url{https://huggingface.co/datasets/zai-org/terminal-bench-2-verified}). Scores are averaged over 5 runs.

CyberGym: We evaluate in Claude Code 2.1.18 (think mode, no web tools) with ($temperature=1.0, top\_p=1.0, max\_new\_tokens=32000$) and a 250-minute timeout per task. Results are single-run Pass@1 over 1,507 tasks.

MCP-Atlas: All models are evaluated in think mode on the 500-task public subset with a 10-minute timeout per task. We use Gemini 3 Pro as the judge model.

$\tau^2$-Bench: We add a small prompt adjustment in Retail and Telecom to avoid failures caused by premature user termination. For Airline, we apply the domain fixes proposed in the Claude Opus 4.5 system card.

Vending-Bench 2: Runs are conducted independently by Andon Labs\footnote{\url{https://andonlabs.com/evals/vending-bench-2}}.

\subsection{Optimized User Simulator for $\tau^2$-Bench}
\label{sec:tau_user_prompt}
We add a small prompt adjustment in Telecom and Retail to avoid failures caused by premature user termination. The optimized prompts are shown in Figure~\ref{fig:tau2-prompt-telecom} and Figure~\ref{fig:tau2-prompt-retail}. These optimized prompts are integrated into the system prompt as follows:

\begin{tcolorbox}[left=0mm,right=0mm,top=0mm,bottom=0mm,boxsep=1mm,arc=0mm,boxrule=0pt, frame empty, breakable]
\begin{lstlisting}
SYSTEM_PROMPT = """"
{global_user_sim_guidelines}

<scenario>
{instructions}
</scenario>

{optimized_user_prompt}
"""".strip()
\end{lstlisting}
\end{tcolorbox}

\begin{tcolorbox}[left=0mm,right=0mm,top=0mm,bottom=0mm,boxsep=1mm,arc=0mm,boxrule=0pt, frame empty, breakable]
\footnotesize
\begin{lstlisting}
# Note:
- Do not generate the '###TRANSFER###' before agent clearly tells "YOU ARE BEING TRANSFERRED TO A HUMAN AGENT. PLEASE HOLD ON.".
    Example:
    Case1:
        - agent: "Would you like me to transfer you to a human agent who can assist you with these options and help get your service restored?"
        - user: "Yes, please transfer me to a human agent.".
    Case2:
        - user: "YOU ARE BEING TRANSFERRED TO A HUMAN AGENT. PLEASE HOLD ON."
        - user: "###TRANSFER###"
\end{lstlisting}
\end{tcolorbox}
\noindent\begin{minipage}{\textwidth}
\captionof{figure}{The optimized user prompt for $\tau^2$-Bench Telecom.}
\label{fig:tau2-prompt-telecom}
\end{minipage}

\begin{tcolorbox}[left=0mm,right=0mm,top=0mm,bottom=0mm,boxsep=1mm,arc=0mm,boxrule=0pt, frame empty, breakable]
\footnotesize
\begin{lstlisting}
# Rules:
- Just generate one line at a time to simulate the user's message.
- Do not give away all the instruction at once. Only provide the information that is necessary for the current step.
- Do not hallucinate information that is not provided in the instruction. Follow these guidelines:
    1. If the agent asks for information NOT in the instruction:
        - Say you don't remember or don't have it
        - Offer alternative information that IS mentioned in the instruction
    2. Examples:
        - If asked for order ID (not in instruction): "Sorry, I don't remember the order ID, can you search for it? My name/email/phone number/zipcode is ..."
        - If asked for email (not in instruction): "I don't have my email handy, but I can give you my name and zip code which are..."
- Do not repeat the exact instruction in the conversation. Instead, use your own words to convey the same information.
- Try to make the conversation as natural as possible, and stick to the personalities in the instruction.
# Constraint Handling:
- Provide requests strictly based on what is explicitly stated in the instruction.
- Do not assume, extend, substitute, or generalize in any form.
- Do not modify or relax constraints on:
- Time / Date
- Budget
- Specific terms (e.g., "same" must not be replaced with "similar")
- Core Rule: Any attribute NOT mentioned in the instruction can be either changed or kept the same
- Examples:
    - If instruction says "exchange red item to blue": Only color must change, other attributes (size, material, etc.) are flexible
    - If instruction says "exchange red item to blue, keep the same size": Both color must change AND size must stay the same
- Exception: Only follow additional constraints when explicitly stated in the instruction
# Domain-Specific Rules:
## For Retail scenarios:
- Focus on product attributes and exchange/return processes as specified in instructions.
- During confirmations: Always respond based strictly on the original instruction, never deviate to match agent's provided options. Restate your requirement from the instruction rather than selecting from agent's choices.
    - Example: If the agent provides specific options (A/B/C) but the instruction states a general requirement (e.g., "same as pending order"), always restate or confirm based on what the instruction says, not by directly selecting from the agent's provided options.
# When NOT to finish the conversation:
- Do not end until you have clearly and completely expressed all your requirements and constraints.
- Do not end until the agent has completed all tasks mentioned in the instruction and verified no operations were missed.
- Do not end if the agent's execution results do not match your expectations or are incorrect/incomplete.
# When you CAN finish the conversation:
- Only when all above conditions are satisfied AND all tasks are completed correctly.
- OR when you have clearly expressed complete requirements but the system explicitly states it cannot complete them due to technical limitations - in this case, accept transfer to human.
# How to finish the conversation:
- If the agent has completed all tasks, generate the '###STOP###' token to end the conversation.
# Note:
- You should carefully check if the agent has completed all tasks mentioned in the instruction before generating '###STOP###'.
\end{lstlisting}
\end{tcolorbox}
\noindent\begin{minipage}{\textwidth}
\captionof{figure}{The optimized user prompt for $\tau^2$-Bench Retail.}
\label{fig:tau2-prompt-retail}
\end{minipage}

\subsection{Evaluation of Real-world Agentic Engineering Experience}

\subsubsection{Frontend Evaluation}

\label{sec:frontend_eval_data_cons}

\paragraph{Data.}
Our dataset encompasses seven distinct frontend scenarios designed to evaluate a model's engineering proficiency across diverse functional domains: Business Management Systems, Web Games, SVG/Canvas Rendering,  Creative Tools \& Editors, Showcase Pages, Forms \& Tables and Data Visualization.

\begin{table}[h]
\centering
\small
\caption{Distribution of frontend application scenarios.}
\begin{tabular}{lp{7cm}rr}
\toprule
\textbf{Category} & \textbf{Description} & \textbf{\# Tasks} & \textbf{\# Checkitems} \\
\midrule
Business Systems & Enterprise/Personal data and process management. & 42 & 167 \\
Web Games & Interaction and entertainment-focused games. & 40 & 163\\
SVG/Canvas & Graphics rendering and interactive visualizations. & 32 & 166\\
Creative Tools & Content creation and online editing tools. & 28 & 160\\
Showcase Pages & Visual expression and information presentation. & 27 & 115\\
Forms \& Tables & Structured data entry and processing. & 26 & 93\\
Data Visualization & Graphical data expression and analysis. & 25 & 85\\
\bottomrule
\end{tabular}

\end{table}

\subparagraph{Data Distribution by Coding Languages}
The benchmark provides full coverage of three mainstream paradigms: Vanilla Web Stack (HTML/CSS/JS), React Component-based Framework, and the Vue 3 + Vite Progressive Solution.
\begin{table}[h]
\centering
\caption{Statistics of technology stacks and evaluation units.}
\begin{tabular}{llcc}
\toprule
\textbf{Category} & \textbf{Description} & \textbf{\# Tasks} & \textbf{\# Checkitems} \\
\midrule
HTML & Vanilla HTML/CSS/JS development. & 113 & 490 \\
React & Component-based framework development. & 58 & 249 \\
Vue & Vue 3 + Vite progressive solution. & 49 & 210 \\
\bottomrule
\end{tabular}
\end{table}

\subparagraph{Data sample}
Each test case is composed of three components: the \textbf{Task}, the \textbf{Checklist}, and a \textbf{Dedicated Environment}. Below is a representative example of a test case:
\begin{tcolorbox}[left=0mm,right=0mm,top=0mm,bottom=0mm,boxsep=1mm,arc=0mm,boxrule=0pt, frame empty, breakable]
\footnotesize
\begin{lstlisting}
    Task: Develop an online drawing tool that includes a brush, an eraser, a white canvas, and a save button.
        The brush color and thickness should be selectable via buttons on the left. Users can draw on the canvas by clicking and dragging the mouse.
        The eraser size should be selectable via buttons on the left. Users can erase content by clicking and dragging the mouse over the canvas.
        Once the drawing is complete, clicking the "Save" button should allow the user to save the image locally.
        Please implement this using the React framework in the current directory.

    Checklist:
        The user can select the brush color and thickness using the left-hand buttons, and drawing is functional via mouse click-and-drag on the canvas.
        The user can select the eraser size using the left-hand buttons, and erasing is functional via mouse click-and-drag on the canvas.
        Upon clicking the "Save" button, the generated image is successfully saved to the local machine.
\end{lstlisting}
\end{tcolorbox}
\noindent\begin{minipage}{\textwidth}
\end{minipage}

\subparagraph{\textbf{Data Construction and Validation}}

We implement a rigorous four-stage pipeline to ensure data quality:
\begin{itemize}[leftmargin=1.5em,itemsep=2pt]
\item Stage 1: Task Synthesis. Tasks are designed by senior frontend experts to ensure they reflect real-world engineering challenges while maintaining a balanced distribution across diverse scenarios and technologies.
\item Stage 2: Checklist Generation and Refinement. We initially employ Claude Sonnet 4.5 to synthesize candidate checklists based on task specifications $T$. These are then meticulously audited and integrated by experts. Through multiple rounds of refinement, we ensure that each check-item is semantically unambiguous, objective, and provides exhaustive coverage of user requirements.
\item Stage 3: Execution-based Correction. We conduct cross-validation between the Agent-as-a-Judge framework and human experts. Any discrepancies in judgment trigger a re-evaluation and correction of the underlying data to eliminate potential noise.
\item Stage 4: Dynamic Benchmark Iteration. To maintain a high level of discriminative power, we iteratively update the test suite by removing trivial tasks that no longer challenge state-of-the-art coding agents. This expert-led curation process culminated in a final set of 220 high-quality frontend coding tasks and their corresponding checklists.
\end{itemize}

\end{document}